\documentclass[letterpaper]{article} 
\usepackage{aaai24}  
\usepackage{times}  
\usepackage{helvet}  
\usepackage{courier}  
\usepackage[hyphens]{url}  
\usepackage{graphicx} 
\usepackage{natbib}  
\usepackage{caption} 
\PassOptionsToPackage{table}{xcolor}

\usepackage{algorithm}
\usepackage{algpseudocode}

\usepackage{times}

\usepackage{soul}
\usepackage{url}
\usepackage[utf8]{inputenc}
\usepackage{makecell}
\usepackage{fdsymbol}
\usepackage{verbatim}
\usepackage{microtype}
\usepackage{amsmath}
\usepackage{tabularx}
\usepackage{graphicx}
\usepackage{multirow}
\usepackage{textcomp}
\usepackage{subcaption}
\usepackage{times}
\usepackage{latexsym}
\usepackage{microtype}
\usepackage{natbib}
\usepackage{booktabs}
\usepackage{appendix}
\usepackage{tikz}
\usepackage{tikz-dependency}
\usepackage{amsmath}
\usepackage{amsfonts}
\usepackage{graphicx}
\usepackage{amsthm}
\usepackage{mathtools}
\usepackage{multirow}
\usepackage{url}
\usepackage{color}
\usepackage{gensymb}
\usepackage{dsfont}
\usepackage{soul}
\definecolor{color1}{rgb}{0.22,0.45,0.70}  
\definecolor{color2}{rgb}{0.45,0.45,0.45}  

\definecolor{NavyBlue}{rgb}{0.1, 0.4, 0.8}
\definecolor{bblue}{HTML}{4F81BD}
\definecolor{rred}{HTML}{C0504D}
\definecolor{ggreen}{HTML}{9BBB59}
\definecolor{ppurple}{HTML}{CCFF99}
\definecolor{darkGreen}{rgb}{0.2,0.5,0.2}
\definecolor{mydarkblue}{rgb}{0,0.08,0.45}
\definecolor{mygreen}{HTML}{CCFFCC}
\definecolor{myred}{HTML}{FFCCCC}
\definecolor{myblue}{HTML}{CCFFFF}
\definecolor{mypp}{HTML}{CCCCFF}

\usepackage{newfloat}
\usepackage{listings}
\usepackage{wasysym}
\usepackage{url}

\DeclareCaptionStyle{ruled}{labelfont=normalfont,labelsep=colon,strut=off} 
\floatstyle{ruled}
\newfloat{listing}{tb}{lst}{}
\floatname{listing}{Listing}

\title{T-SciQ: Teaching Multimodal Chain-of-Thought Reasoning via Mixed Large Language Model Signals for Science Question Answering}
\author{
    Lei Wang\textsuperscript{\rm 1},
    Yi Hu\textsuperscript{\rm 2},
    Jiabang He\textsuperscript{\rm 2},
    Xing Xu\textsuperscript{\rm 2},
    Ning Liu\textsuperscript{\rm 3},
    Hui Liu\textsuperscript{\rm 4},
    Heng Tao Shen\textsuperscript{\rm 2}
}
\affiliations{
    \textsuperscript{\rm 1} Singapore Management University \,\,\,
    \textsuperscript{\rm 2} University of Electronic Science and Technology of China \\
    \textsuperscript{\rm 3} Beijing Forestry University \,\,\,
    \textsuperscript{\rm 4} Beijing Rongda Technology Co., Ltd. \\

    lei.wang.2019@phdcs.smu.edu.sg, yihu0118@gmail.com, JiaBangH@outlook.com \\ xing.xu@uestc.edu.cn, liuning0928@bjfu.edu.cn, ryuki122382@gmail.com, shenhengtao@hotmail.com
}

\usepackage{bibentry}

\begin{document}

\maketitle

\begin{abstract}
Large Language Models (LLMs) have recently demonstrated exceptional performance in various Natural Language Processing (NLP) tasks. They have also shown the ability to perform chain-of-thought (CoT) reasoning to solve complex problems. Recent studies have explored CoT reasoning in complex multimodal scenarios, such as the science question answering task, by fine-tuning multimodal models with high-quality human-annotated CoT rationales. However, collecting high-quality COT rationales is usually time-consuming and costly. Besides, the annotated rationales are hardly accurate due to the external essential information missed.
To address these issues, we propose a novel method termed \emph{T-SciQ} that aims at teaching science question answering with LLM signals. The T-SciQ approach generates high-quality CoT rationales as teaching signals and is advanced to train much smaller models to perform CoT reasoning in complex modalities. Additionally, we introduce a novel data mixing strategy to produce more effective teaching data samples for simple and complex science question answer problems.
Extensive experimental results show that our T-SciQ method achieves a new state-of-the-art performance on the ScienceQA benchmark, with an accuracy of 96.18\%. Moreover, our approach outperforms the most powerful fine-tuned baseline by 4.5\%. The code is publicly available at \url{https://github.com/T-SciQ/T-SciQ}.
\end{abstract}

\section{Introduction}

\begin{figure}[t]
    \centering
    \includegraphics[width=0.94\linewidth]{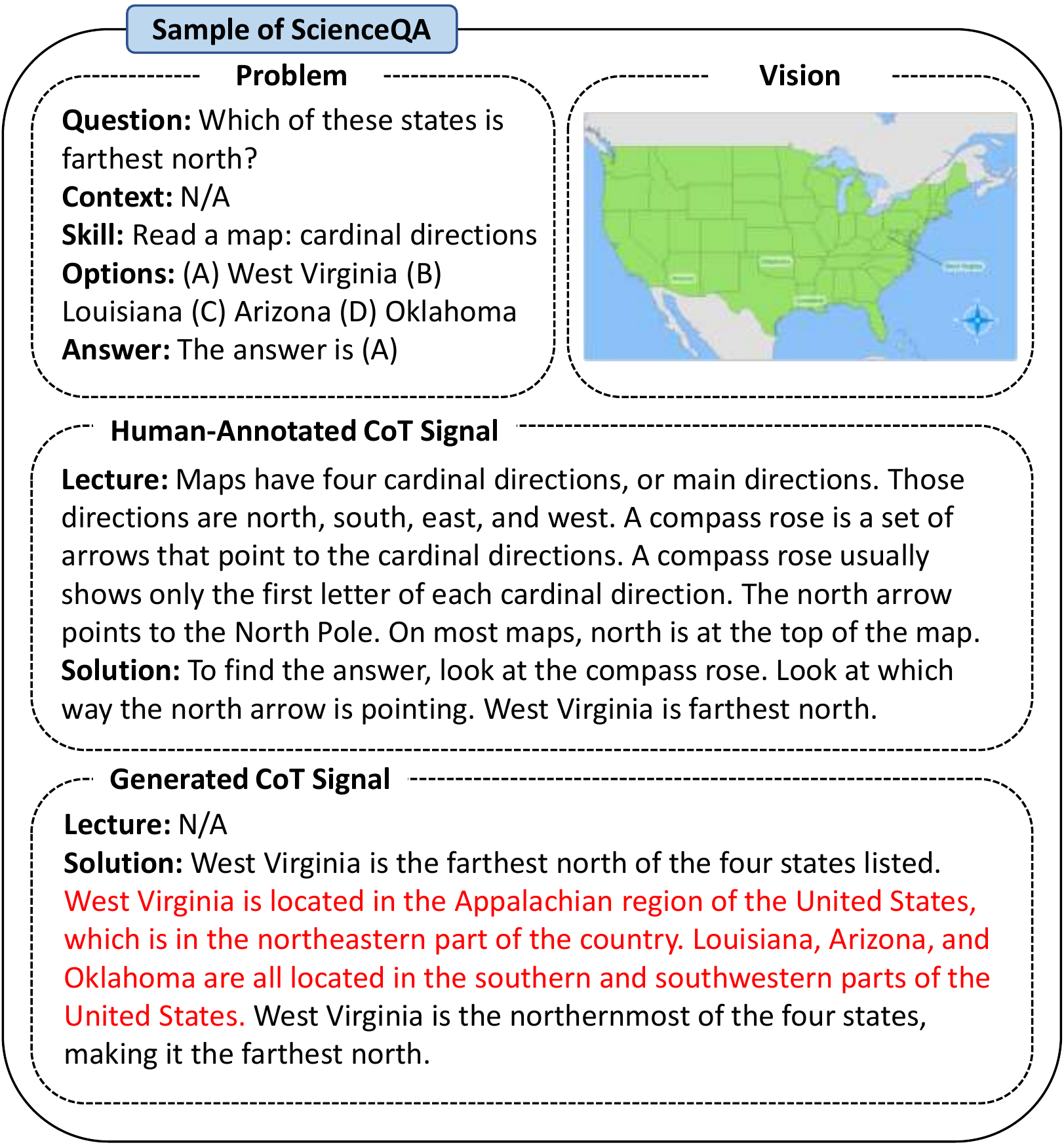}
    \vspace{-5pt}
    \caption{The input of a ScienceQA data example includes a question, context, image, skill, and options. Annotations include the ground truth answer and CoT rationale (lecture and solution). Compared to annotated CoT, LLM-generated CoT includes greater amounts of essential external knowledge.}
    \label{fig:intro}
    \vspace{-15pt}
\end{figure}

Scientific problem solving has recently been employed to evaluate the multi-hop reasoning capability and interpretability of AI systems~\cite{kembhavi2017you, sampat2020visuo, dalvi2021explaining}. However, these datasets~\cite{kembhavi2017you, jansen2018worldtree} suffer from limited scale. To address this issue, \citet{lu2022learn} introduces a large-scale science question-answering dataset across broad topics and skills called ScienceQA. This dataset consists of 21,208 multimodal data examples associated with questions, context, images, options, lectures, and explanations. An example is shown in Figure~\ref{fig:intro}, illustrating that a model must comprehend multimodal inputs and incorporate external knowledge to answer scientific questions.

Recently, Large Language Models (LLMs) have shown exceptional performance in various Natural Language Processing (NLP) tasks \cite{brown2020language, thoppilan2022lamda}. Specifically, they have demonstrated the chain-of-thought (CoT) ability to solve complex reasoning problems by using a few demonstration examples without additional training \citep{cot_wei,kojima2022large,zhang2022automatic}. 
However, the existing research on CoT reasoning is mainly limited to the language modality \citep{wang2022rationale,zhou2022least,lu2022dynamic,fu2022complexity}, with little attention paid to multimodal scenarios, such as science question answering.
To address this issue, a common approach is to use caption models to translate visual information into the language modality and prompt LLMs to perform CoT reasoning~\cite{lu2022learn}. However, the use of caption generation models in scientific problems may result in significant information loss when meeting highly complex images. 
To overcome this issue, \citet{zhang2023multimodal} proposed a framework called Multimodal-CoT that models both language and visual modalities into a two-stage fine-tuning process, which separates rationale generation and answer inference.

The Multimodal-CoT method has a significant disadvantage because it relies on the human-annotated CoT rationale to fine-tune the model. While incorporating human-annotated CoT signals is helpful for training models to facilitate CoT reasoning ability, it has two fundamental limitations.
First, the human annotation of CoT reasoning is time-consuming~\cite{nye2021show, cobbe2021training}, particularly for complex tasks like ScienceQA, which necessitates extensive expert knowledge to create a reasoning process for the answer.
Second, as shown in Figure~\ref{fig:intro}, the annotated rationale may lack essential external information to derive the final answer due to the limited expertise of human annotators.

To address these issues, we propose a novel approach named \emph{T-SciQ} to solve the ScienceQA task. 
The proposed T-SciQ framework consists of three stages: generating teaching data, mixing teaching data, and fine-tuning.
For teaching data generation, we use a simple zero-shot instruction and a hint of the correct answer to generate a CoT rationale for a QA data example to obtain a QA-CoT sample. 
Although the model taught by QA-CoT samples excels at tackling simple problems, it still struggles with highly complicated problems.
To overcome this challenge, we follow the zero-shot plan-and-solve prompting~\cite{wang2023plan} to generate plan-based CoT (PCoT) rationales, which decompose complex problems into simpler subproblems to solve, to obtain QA-PCoT teaching samples.
To combine the strengths of both teaching signals, we create a new teaching dataset called T-SciQ by mixing QA-CoT and QA-PCoT datasets.
Specifically, 
we use the validation set to determine whether the PCoT teaching signal or CoT teaching signal is more appropriate for each data example in a given skill.
Then, we fine-tune the student model with teaching data. We follow the Multimodal-CoT~\citep{zhang2023multimodal} to build our student model, which consists of two-stage: rationale generation teaching and answer inference teaching. 
During inference, the model trained in the first stage generates rationales for the test data. The generated rationales are subsequently used in the second stage to infer answers.

Experiment results on the ScienceQA benchmark show that our method surpasses the previous state-of-the-art by a large margin.
Specifically, the student model taught by T-SciQ teaching samples outperforms the most powerful fine-tuned baseline by 4.5\%, the strongest instruction-tuning based multimodal baseline by 5.26\%, the best GPT-4 based few-shot baseline by 9.64\%, and human performance by 7.78\%.
To demonstrate the versatility of our teaching approach, we additionally conduct experiments to compare Reason-Teacher~\cite{ho2022large_teachers} on six reasoning tasks.
Our main contributions are summarized as follows: 1) We propose a novel framework for generating high-quality CoT rationale and training student models to perform CoT reasoning for the ScienceQA task; 2) We introduce a data mixing strategy to produce effective teaching data samples for simple and complex problems; 3) Our method achieves a new state-of-the-art performance on the ScienceQA benchmark, surpassing all previous models by a large margin.

\begin{figure*}[!htb]
    \centering
    \includegraphics[width=0.99\linewidth]{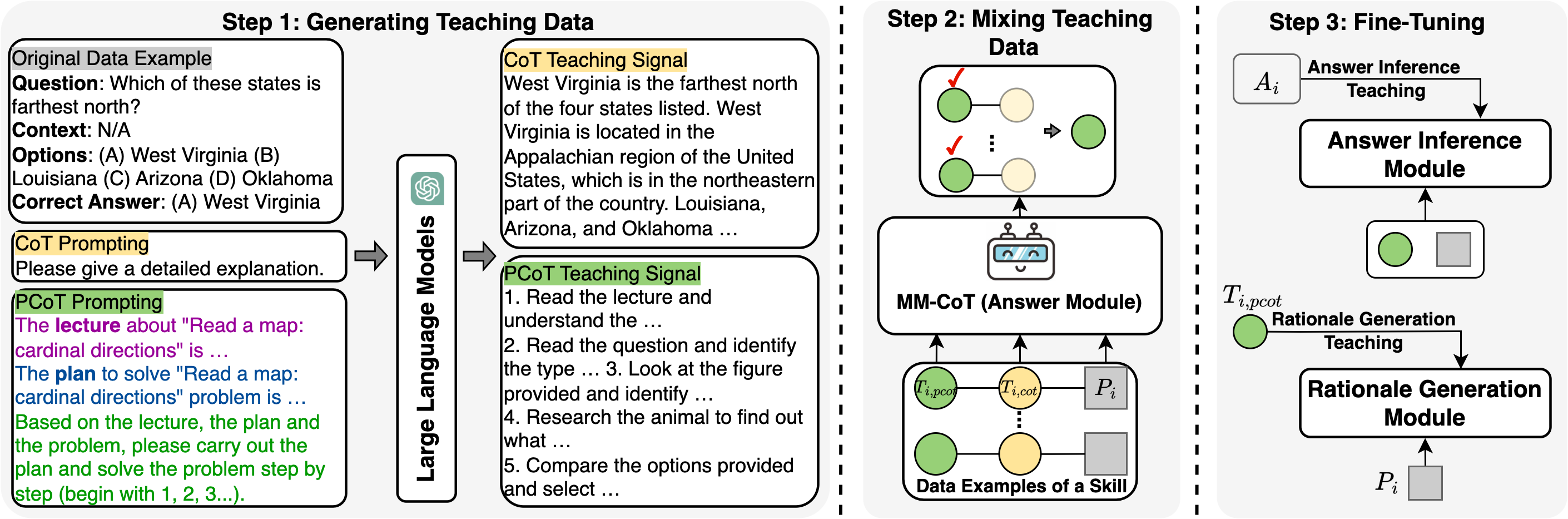}
    \vspace{-5pt}
    \caption{
        Key steps of our T-SciQ approach. T-SciQ consists of three stages: (i) generating teaching data; (ii) mixing teaching data; and (iii) fine-tuning. 
    }
    \label{fig:method}
    \vspace{-15pt}
\end{figure*}

\section{Related Work}
\subsubsection{Chain-of-Thought Prompting.}
Recently, to solve complex reasoning tasks, 
~\citet{wei2022chain} propose CoT prompting by prompting large language models to generate intermediate reasoning processes before reaching the final answer. Subsequently, a lot of work has been proposed to further improve CoT prompting from different aspects, including improving the quality of demonstrations~\citep{rubin2021learning, zhang2022automatic, fu2022complexity, lu2022dynamic, he2023icl} and improving the quality of reasoning chains~\citep{zhou2022least, khot2022decomposed, chen2022program, wang2022self, wang2022rationale, li2022advance, tian2023r}.
Zero-shot CoT~\citep{kojima2022large} elicited reasoning step by appending a prompt like ``\textit{Let's think step by step}'' to the test question. 
Iterative Prompting~\citep{wang2022iteratively} dynamically synthesizes prompts conditioned on the current step's contexts. 
PoT Prompting~\citep{chen2022program} writes a program as a rationale and invoked the reasoning ability of LLMs by executing the generated program. 
Chameleon~\citep{lu2023chameleon} proposed a plug-and-play compositional reasoning framework to utilize multiple modules to obtain high quality prompting. 
Our work mainly focuses on mixing different teaching CoT rationales for different problems.

\noindent\textbf{LLMs as Teachers.}
In recent studies, CoT reasoning is elicited in small models using fine-tuned language models. 
\citet{magister2022teaching} benefit smaller models through CoT distillation.
\citet{huang2022large} show that LLMs can enhance reasoning using self-generated solutions from unlabeled data.
\citet{ho2022large_teachers} propose Fine-tune-CoT to leverage the capabilities of LLMs to generate reasoning samples and teach smaller models via fine-tuning. 
Distilling step-by-step~\cite{hsieh2023distilling} improves small model performance using LLM rationales with less data.
Multimodal-CoT~\citep{zhang2023multimodal} uses two-stage fine-tuning with annotated CoT rationales and visual features to achieve state-of-the-art results on the ScienceQA benchmark.
Our work exploits generating two types of teaching data from LLMs and mixing teaching data.
We discover that this simple method highly improves student performance in complex multi-modality tasks, which has not yet been recognized in previous studies on fine-tuning with CoT reasoning~\cite{hsieh2023distilling, ho2022large_teachers, huang2022large, magister2022teaching, fu2023specializing, hu2023llm}. 

\section{Our T-SciQ Approach}



\subsection{Overview}
This section presents the proposed fine-tuning strategy T-SciQ, which utilizes a LLM named SciTeacher to generate teaching data and improve the performance of a smaller student model (SciStudent) by generated teaching data. The proposed T-SciQ strategy comprises three components: generating teaching data, mixing teaching data, and fine-tuning, as depicted in Figure~\ref{fig:method}.
To generate the teaching data, we leverage SciTeacher to produce CoT rationales to obtain Question-Answer-CoT (QA-CoT) samples, and planning-based CoT rationale (PCoT) to obtain Question-Answer-PCoT (QA-PCoT) samples. 
To combine the strengths of both datasets, we create a new teaching dataset called T-SciQ by mixing QA-CoT and QA-PCoT datasets.
Specifically, we use the validation set to determine whether the PCoT teaching signal or CoT teaching signal is more appropriate for each data example in a given skill.
We then use T-SciQ teaching samples to fine-tune the smaller student models.
In the following, we provide a detailed description of these three components.

\subsection{Generating Teaching Data}

We produce two types of data samples for teaching: QA-CoT sample with a generated CoT rationale and QA-PCoT sample equipped with a generated PCoT rationale.

\paragraph{QA-CoT Sample Generation.} Although using human-annotated CoT signals is valuable for training models to elicit CoT reasoning ability, it has two inherent limitations: time-consuming and lack of external essential information due to human annotators' restricted expertise.

To address these issues, we introduce a zero-shot prompting to generate high-quality CoT rationales from LLMs. We achieve this by converting the input training data example $X$ into a prompt, utilizing a straightforward template that reads as follows:
``Question: \texttt{[$X_q$]}. Context: \texttt{[$X_c$]}. Options: \texttt{[$X_o$]}. Correct Answer: \texttt{[$A$]}. \texttt{[$Instruct$]}''.
Here, the \texttt{[$X_q$]} slot is for the input question, the \texttt{[$X_c$]} slot is for the input context, the \texttt{[$X_o$]} slot contains the possible options, the \texttt{[$A$]} slot is for the correct answer that can work as a hint to guide LLMs to generate a more reliable rationale, and the \texttt{[$Instruct$]} slot contains instructions, i.e., ``\textit{Please give me a detailed explanation.}'', to guide LLMs to perform the task. Note that the context may not be included for some data examples, in which case the context slot is replaced with ``N/A''. 
Subsequently, we feed the filled prompt to LLMs to output a reasoning process for a given training data example to obtain QA-CoT data $D_{\text{QA-CoT}}$.

\paragraph{QA-PCoT Sample Generation.}
Although using QA-CoT samples can address issues of human-annotated CoT, addressing highly complex problems remains a challenge. 
To overcome this challenge and obtain appropriate teaching CoT rationale, we introduce a 3-step zero-shot prompting to decompose complex problems into simpler subproblems.

\noindent \textbf{Step 1: Lecture Generation.} 
The lecture template used to generate a lecture for a particular skill is formulated as follows: ``Skill: \texttt{[$S$]}. QA pairs: \texttt{[$X_q, A$]} ... \texttt{[$Instruct$]}.'' In this prompt, \texttt{[$Instruct$]} is as follows: ``\textit{based on the problems above, please give a general lecture on the \texttt{[$S$]} type of question in one sentence.}''.
Note that many QA examples need the same skill to be solved.

\noindent \textbf{Step 2: Plan Generation.}
The template used to generate a plan for a specific skill based on the generated lecture is formulated as follows: ``Skill: \texttt{[$S$]}. Lecture: \texttt{[$L$]}. QA pairs: \texttt{[$X_q, A$]} ... \texttt{[$Instruct$]}.''. In this prompt, \texttt{[$Instruct$]} is written as follows: ``\textit{Based on the lecture above and these problems, let’s understand these problems and devise a general and brief plan step by step to solve these problems (begin with 1, 2, 3...)}''.


\noindent \textbf{Step 3: Rationale Generation.} 
The lecture and plan generated by the first two prompts are used to generate a plan-based CoT rationale for each training example. The rationale generation template is formulated as follows:
``Skill: \texttt{[$S$]}. Lecture: \texttt{[$L$]}. Plan: \texttt{[$P$]}. QA pair: \texttt{[$X_q, A$]}.  \texttt{[$Instruct$]}.''. 
In this prompt, \texttt{[$Instruct$]} is written as follows: ``Based on the lecture, the plan and the problem, please carry out the plan and solve the problem step by step (begin with 1, 2, 3...)''.
Examples of this three-step prompting can be found in the supplementary material.

\subsection{Mixing Teaching Data}
The QA-PCoT dataset is effective for teaching problem-solving skills for complex problems, while simpler problems don't require decomposition. In contrast, the QA-CoT dataset is suitable for teaching problem-solving skills for simple problems. To combine the strengths of both datasets, we create a new teaching dataset called T-SciQ by mixing QA-CoT and QA-PCoT datasets.
We introduce a new approach that uses the validation set to determine whether the PCoT teaching signal or CoT teaching signal is more appropriate for a data example in a given skill.

Given a ScienceQA problem $P_i$ with the language input $X_{i,\textrm{la}}$ and the visual input $X_{i,\textrm{v}}$, our objective is to let an answer generation model $F^s_a$ help identify the optimal teaching signal $T_{i,k}$ from the possible choices $T_{i}$, i.e., CoT teaching signal $T_{i,\text{cot}}$ or PCoT teaching signal $T_{i,\text{pcot}}$, thereby maximizing the answer accuracy of the validation set.
The answer generation module $F^s_a$ is similar to the one described in Multimodal-CoT~\cite{zhang2023multimodal}. 
The generated answer $\hat{A}_i$ is produced by $F^s_a(X_{i, \text{la}}, X_{i,\text{v}}, T_{i,k})$, and the number of errors is obtained by comparing the generated answer $\hat{A}_i$ and the label $A_i$. If the number of errors for validation samples with PCoT in a skill is lower than that of validation samples with CoT in a skill, we select PCoT rationale as the teaching rationale for all training data examples in this skill. Otherwise, we select CoT rationale. The obtained teaching samples are then used to fine-tune the student model.
To train the answer generation module, we utilize a subset of training data examples, each of which is associated with the human-annotated teaching signal from the original ScienceQA dataset.

\subsection{Fine-Tuning}

\begin{table*}[t]
\centering
\small
\renewcommand\tabcolsep{6.0pt} 
\caption{Main results (\%) on the test set of ScienceQA. There are totally 8 classes of questions, namely natural science (\textbf{NAT}), social science (\textbf{SOC}), language science (\textbf{LAN}), text context (\textbf{TXT}), image context (\textbf{IMG}), no context (\textbf{NO}), grades 1-6 (\textbf{G1-6}), and grades 7-12 (\textbf{G7-12}). The best results are boldfaced. The improvements are shown in {\color{blue}blue}.}
\vspace{-10pt}
{
\begin{tabular}{l|r|cccccccc|l} 
\toprule
 Model  & Size & NAT & SOC & LAN & TXT & IMG & NO & G1-6 & G7-12 & ~Avg \\
\midrule
Human & - & 90.23  & 84.97 & 87.48 & 89.60 & 87.50 & {88.10} & 91.59 & 82.42 & 88.40 \\
 \midrule
 MCAN \citep{yu2019deep} & 95M & 56.08 & 46.23 & 58.09 & 59.43 & 51.17 & {55.40} & 51.65 & 59.72 & 54.54 \\
 Top-Down \citep{anderson2018bottom} & 70M & 59.50 & 54.33 & 61.82 & 62.90 & 54.88 & {59.79} & 57.27 & 62.16 & 59.02 \\
 BAN \citep{kim2018bilinear}  & 112M & 60.88 & 46.57 & 66.64 & 62.61 & 52.60 & {{65.51}} & 56.83 & 63.94 & 59.37 \\
 DFAF \citep{gao2019dynamic}  & 74M & 64.03 & 48.82 & 63.55 & 65.88 & 54.49 & {64.11} & 57.12 & 67.17 & 60.72 \\
 ViLT \citep{kim2021vilt}  & 113M & 60.48 & 63.89 & 60.27 & 63.20 & 61.38 & {57.00} & 60.72 & 61.90 & 61.14 \\
 Patch-TRM \citep{lu2021iconqa} & 90M & {65.19} & 46.79 & {65.55} & {66.96} & 55.28 & {64.95} & 58.04 & {67.50} & 61.42 \\
 VisualBERT \citep{li2019visualbert}  & 111M & 59.33 & {69.18} & 61.18 & 62.71 & {62.17} & {58.54} & {62.96} & 59.92 & {61.87} \\
 UnifiedQA$_\texttt{Base}$ \citep{khashabi2020unifiedqa}& 223M &68.16 & 69.18 & 74.91 & 63.78 & 61.38 & {77.84} & 72.98 & 65.00 & 70.12 \\
 \midrule
LLaMa-Adapter~\citep{zhang2023llama_adapter}  & $>$7B & 84.37 & 88.30 & 84.36 & 83.72 & 80.32 & 86.90 & 85.83 & 84.05 & 85.19 \\
LLaVA~\citep{liu2023visual_instruction} & $>$7B & {90.36} & {95.95} & {88.00} & {89.49} & {88.00} & {{90.66}} & {90.93} & {90.90} & {90.92} \\
\midrule
GPT-3.5 \citep{chen2020big}  & $>$175B & 74.64 & 69.74 & 76.00 & 74.44 & 67.28 & {77.42} & 76.80 & 68.89 & 73.97 \\
GPT-3.5 w/ CoT \citep{lu2022learn}  & $>$175B& {75.44} & {70.87} & {78.09} & {74.68} & {67.43} & {{79.93}} & {78.23} & {69.68} & {75.17} \\
ChatGPT w/ CoT~\citep{lu2023chameleon} & $>$175B & 78.82 & 70.98 & 83.18 & 77.37 & 67.92 & 86.13 & 80.72 & 74.03 & 78.31 \\
GPT-4 w/ CoT~\citep{lu2023chameleon} & $>$175B& {84.06} & {73.45} & {87.36} & {81.87} & {70.75} & {{90.73}} & {84.69} & {79.10} & {82.69} \\
Chameleon~\citep{lu2023chameleon} & $>$175B & {89.83} & {74.13} & {89.82} & {88.27} & {77.64} & {{92.13}} & {88.03} & {83.72} & {86.54} \\
\midrule
 UnifiedQA-CoT$_\texttt{Base}$ \citep{lu2022learn} & 223M & {71.00} & {76.04} & {78.91} & {66.42} & {66.53} & {81.81} & {77.06} & 68.82 & {74.11}  \\
 \rowcolor{gray!25} \textbf{UnifiedQA-T-SciQ$_\texttt{Base}$ (Ours)} & 223M & 76.56 & 88.99 & 80.45 & 72.90 & 73.84 & 83.47 & 81.09 & 75.19 & 79.41 \\
\rowcolor{gray!25}Improvement & - & \textcolor{blue}{+5.56} & \textcolor{blue}{+12.95} & \textcolor{blue}{+1.54} & \textcolor{blue}{+6.48} & \textcolor{blue}{+7.31} & \textcolor{blue}{+1.66} & \textcolor{blue}{+4.03} & \textcolor{blue}{+6.37} & \textcolor{blue}{+5.30} \\
  \midrule
 Mutimodal-CoT$_\texttt{Base}$  \citep{zhang2023multimodal} & 223M & {87.52} & {77.17} & {85.82} & {87.88} & {82.90} & {86.83} & {84.65} & {85.37} & {84.91} \\
 \rowcolor{gray!25} \textbf{Mutimodal-T-SciQ$_\texttt{Base}$ (Ours)}  & 223M & 91.52 & 91.45  & 92.45 & 91.94 & 90.33 & 92.26 & 92.11 & 91.10 & 91.75 \\
 \rowcolor{gray!25}Improvement & - & \textcolor{blue}{+4.00} & \textcolor{blue}{+14.28} & \textcolor{blue}{+6.63} & \textcolor{blue}{+4.06} & \textcolor{blue}{+7.43} & \textcolor{blue}{+5.43} & \textcolor{blue}{+7.46} & \textcolor{blue}{+5.73} & \textcolor{blue}{+6.84} \\
\midrule
Mutimodal-CoT$_\texttt{Large}$ \citep{zhang2023multimodal} & 738M & 95.91 & 82.00 & 90.82 & 95.26 & 88.80 & 92.89 & 92.44 & 90.31 & 91.68 \\
\rowcolor{gray!25} \textbf{Mutimodal-T-SciQ$_\texttt{Large}$ (Ours)}  & 738M & \textbf{96.89} & \textbf{95.16} & \textbf{95.55} & \textbf{96.53} & \textbf{94.70} & \textbf{96.79} & \textbf{96.44} & \textbf{95.72} & \textbf{96.18} \\
\rowcolor{gray!25}Improvement & - & \textcolor{blue}{+0.98} & \textcolor{blue}{+13.16} & \textcolor{blue}{+4.73} & \textcolor{blue}{+1.27} & \textcolor{blue}{+5.90} & \textcolor{blue}{+3.90} & \textcolor{blue}{+4.00} & \textcolor{blue}{+5.41} & \textcolor{blue}{+4.50} \\
 \bottomrule
\end{tabular}
}
\vspace{-10pt}
\label{tab:main_results}
\end{table*}

Our teaching follows the Multimodal-CoT~\citep{zhang2023multimodal} two-stage fine-tuning framework: rationale generation teaching and answer inference teaching. 

\paragraph{Rationale Generation Teaching.} 
In this stage, the rationale generation model $F_r(P_i)$ is trained to predict the teaching signal $T_i$ for a given problem $P_i$, where $T_i$ either be CoT rationale or PCoT rationale. The input of $F_r(P_i)$ consists of $X_{i,\textrm{la}}^{1}$ and $X_{i,\textrm{v}}$, where $X_{i,\textrm{la}}^{1}$ represents the language input and $X_{i,\textrm{v}}$ represents the visual input.
Formally, the probability of generating rationale $T_i$ can be formulated as follows:
\begin{equation}
p(T_i|X_{i,\textrm{la}^1},X_{i,\textrm{v}}) = \prod_{j=1}^{N_{T_i}} p_{\theta_r}\left(T_{i,j} \mid X_{i,\textrm{la}}^1, X_{i,\textrm{v}}, T_{i,<j}\right),
\end{equation}
where $\theta_r$ represents learnable parameters of the rationale generation model $F_r$ and $N_{T_i}$ is the length of $T_{i}$. 

\paragraph{Answer Inference Teaching.} In the second stage, we construct the language input $X_{i,\textrm{la}}^{2}$ by appending the teaching rationale $T_i$ to the original language input $X_{i,\textrm{la}}^{1}$. The new input $X'_i$ is then fed to the answer inference model to infer the final answer $A_i = F_a(X'_i)$, where $X'_i=\{X_{i,\textrm{la}}^{2}, X_{i,\textrm{v}}\}$.
Formally, the probability of generating answer $A_i$ can be formulated as follows:
\begin{equation}
p(A_i|X^2_{i,\textrm{la}},X_{i,\textrm{v}}) = \prod_{j=1}^{N_{A_i}} p_{\theta_a}\left(A_{i} \mid X^2_{i,\textrm{la}}, X_{i,\textrm{v}}, A_{i,<j}\right),
\end{equation}
where $\theta_a$ represents learnable parameters in the answer inference teaching stage. 

\paragraph{Model Architecture}
We utilize the Multimodal-CoT~\citep{zhang2023multimodal} model architecture as our default, which employs a Transformer model~\cite{transformer} for encoding language and a vision Transformer for encoding visual information. The gated fusion mechanism, proposed in \citep{li2022vision}, is used to effectively integrate the language and vision representations. Finally, a Transformer decoder is used to generate the target output. Note that rationale generation and answer inference share the same model but differ in the input and output. 



\section{Experiment}

\begin{table}[t]
\centering
\small
\renewcommand\tabcolsep{8pt} 
\caption{Ablation study of the impact of different signals provided by LLMs across all topics. }
\vspace{-5pt}
{
\begin{tabular}{l|c} 
\toprule
 Model & ~Avg \\
\midrule
Multimodal-T-SciQ$_\texttt{Base}$ (Mixing) &  91.75 \\
Multimodal-T-SciQ$_\texttt{Base}$ only w/ QA-CoT & 85.99 \\
Multimodal-T-SciQ$_\texttt{Base}$ only w/ QA-PCoT& 88.56 \\
Mutimodal-CoT$_\texttt{Base}$ &  {84.91} \\
\midrule
Multimodal-T-SciQ$_\texttt{Large}$ (Mixing) & 96.18 \\
Multimodal-T-SciQ$_\texttt{Large}$ only w/ QA-CoT &  93.44 \\
Multimodal-T-SciQ$_\texttt{Large}$ only w/ QA-PCoT & 94.11 \\
 Mutimodal-CoT$_\texttt{Large}$ & 91.68 \\
\bottomrule
\end{tabular}
}
\vspace{-15pt}
\label{tab:ablation_main}
\end{table}

\subsection{Experimental Setup}
\noindent\textbf{Dataset.} We evaluate our proposed method on the \textbf{ScienceQA}~\citep{lu2022learn} dataset, a latest multimodal multiple-choice science question dataset comprising 21,208 examples. 
ScienceQA encompasses a wide range of topics across three distinct subjects: natural science, social science, and language science. The dataset comprises 26 topics, 127 categories, and 379 skills that are relevant to these three subjects.
We employ the official split provided by ScienceQA, which divides the dataset into training, validation, and test sets with a ratio of $3$:$1$:$1$, i.e., $12,726$, $4,241$, and $4,241$ examples, respectively.
The dataset includes annotated reasoning chains for each data example.
In this work, we extract our training signals from large language models instead of using human annotated signals.

\noindent\textbf{Baselines.} We provide a comparison of our proposed method with extensive baseline methods.
Specifically, we have several early VQA models, including \texttt{MCAN} \citep{yu2019deep}, \texttt{Top-Down} \citep{anderson2018bottom}, \texttt{BAN} \citep{kim2018bilinear}, \texttt{DFAF} \citep{gao2019dynamic}. These VQA baselines use the question, context, and answer choices as textual input and the image as the visual input. They predict a score distribution over the answer candidates using a linear classifier.
In addition, we include pre-trained text-to-text and multimodal models such as \texttt{ViLT}~\citep{kim2021vilt}, \texttt{Patch-TRM}~\citep{lu2021iconqa}, and \texttt{VisualBERT}~\citep{li2019visualbert}, \texttt{UnifiedQA}~\citep{khashabi2020unifiedqa}, \texttt{MM-COT}~\citep{zhang2023multimodal}. These methods use pre-trained models as backbone models and incorporate additional modules to handle multimodal signals if necessary.
We also include recent LLM-based multimodal fine-tuned baselines such as \texttt{LLaMa-Adapter}~\citep{zhang2023llama_adapter} and \texttt{LLaVA}~\citep{liu2023visual_instruction}. They use a strong open-access LLM such as \texttt{LLaMa}~\citep{touvron2023llama} as the base model and incorporate a vision module to model visual information.
We also include widely-used in-context learning baselines: the chain of thought (CoT) prompting~\citep{cot_wei}, where each in-context demonstration example comprises the input question and output annotated reasoning process. We compare to the CoT baselines over different API-based OpenAI LLMs~\cite{openai-chatgpt-2022, openai-gpt4-2023}, such as 
GPT-3.5 (\texttt{GPT-3.5 w/ COT}), 
ChatGPT (\texttt{ChatGPT w/ COT}),  GPT-4 (\texttt{GPT-4 w/ COT}), and \texttt{Chameleon}~\citep{lu2023chameleon}. 
Additionally, we also compare to the standard few-shot prompting approach using GPT-3.5 (\texttt{GPT-3.5}).


\noindent\textbf{Evaluation Metrics.} As ScienceQA is a benchmark for multiple-choice question answering, the \textit{accuracy} of the answer is evaluated by comparing the ground truth option with the final prediction generated by the evaluated model. 

\noindent\textbf{Implementation Details.} By default, we utilize the GPT-3.5 of text-davinci-003 version as the teacher model for our approach unless otherwise specified. To validate the generalizability of our method, we experiment with three distinct student models, namely UnifiedQA$_\texttt{Base}$ w/ CoT~\citep{lu2022learn}, Mutimodal-CoT$_\texttt{Base}$\citep{lu2022learn}, and Mutimodal-CoT$_\texttt{Large}$\citep{zhang2023multimodal}. These models are chosen due to their strong performances achieved by fine-tuning with annotated reasoning signals. To ensure fairness of comparison and effectiveness of our proposed method, we only replace the training signals generated by our approach with annotated signals while maintaining the same settings as the original paper.
These student models are 200$\times$ smaller than their teacher models.

\subsection{Main Results}


\subsubsection{T-SciQ \emph{v.s.} Baselines.}
Table~\ref{tab:main_results} details the performance accuracy of baselines and student models trained using the proposed T-SciQ signals. 
Mutimodal-T-SciQ$_\texttt{Large}$, which is the model architecture of Mutimodal-CoT$_\texttt{Large}$ fine-tuned with mixed teacher signals, attains an accuracy of 96.18\% and consistently outperforms all state-of-the-art methods by a large margin for all topics across all subjects. 
Specifically, Mutimodal-T-SciQ$_\texttt{Large}$ outperforms the most powerful fine-tuning baseline, Mutimodal-CoT$_\texttt{Large}$, which is trained by annotated chain-of-thought signals, by 4.5\% (91.68\% $\rightarrow$ 96.18\%), the strongest instruction-tuning based multimodal baseline, LLaVa, by 5.26\% (90.92\% $\rightarrow$ 96.18\%), the best GPT-4 based few-shot baseline, Chameleon, by 9.64\% (86.54\% $\rightarrow$ 96.18\%), and human performance by 7.78\% (88.40\% $\rightarrow$ 96.18\%). 
This significant improvement of our proposed method suggests that higher-quality teaching signals of planning and reasoning  provided by LLMs elicit better planning and chain-of-thought reasoning ability in student models smaller than 1B.

\subsubsection{T-SciQ with Different Base Student Models.} 
Instead of only using the model architecture of Mutimodal-CoT$_\texttt{Large}$ as the base student model, we evaluate different base student models fine-tuned with mixed teaching signals: the variant UnifiedQA-T-SciQ$_\texttt{Base}$  and Mutimodal-T-SciQ$_\texttt{Base}$. 
The relative performance ranking between the base student model with annotated CoT signals and the one with mixing teacher signals remains unchanged.
Specifically, UnifiedQA-T-SciQ$_\texttt{Base}$ outperforms UnifiedQA$_\texttt{Base}$ w/ CoT by 5.3\% (74.11\% $\rightarrow$ 79.41\%), and Mutimodal-T-SciQ$_\texttt{Base}$ outperforms  Mutimodal-CoT$_\texttt{Base}$ by 6.84\% (84.91\% $\rightarrow$ 91.75\%).
T-SciQ still achieves the best performance with different base student models. These encouraging results indicate the generalizability of the proposed teaching signals.

\begin{table}[t]
\centering
\small
\renewcommand\tabcolsep{3pt} 
\caption{Accuracy (\%) of Mutimodal-T-SciQ$_\texttt{Base}$ using different visual features.}
\vspace{-10pt}
\resizebox{0.3\textwidth}{!}{
\begin{tabular}{l|ccc} 
\toprule
 \multirow{2}{*}{Method} & \multicolumn{3}{c}{T-SciQ} \\
  & QA-CoT & QA-PCoT & T-SciQ \\
\midrule
Language Only & 84.44 & 85.38 & 87.24 \\
\midrule
\quad w/ CLIP & 86.18 & 87.41 & 90.90 \\
\quad w/ DETR & 85.99 & 88.56 & 91.75 \\
\quad w/ ResNet & 86.06 & 87.69 & 91.44 \\
\bottomrule
\end{tabular}
}
\vspace{-5pt}
\label{tab:ablation_vision}
\end{table}

\begin{figure}[t]
    \centering
    \begin{subfigure}{0.23\textwidth}
    \centering
    \includegraphics[width=1.0\linewidth]{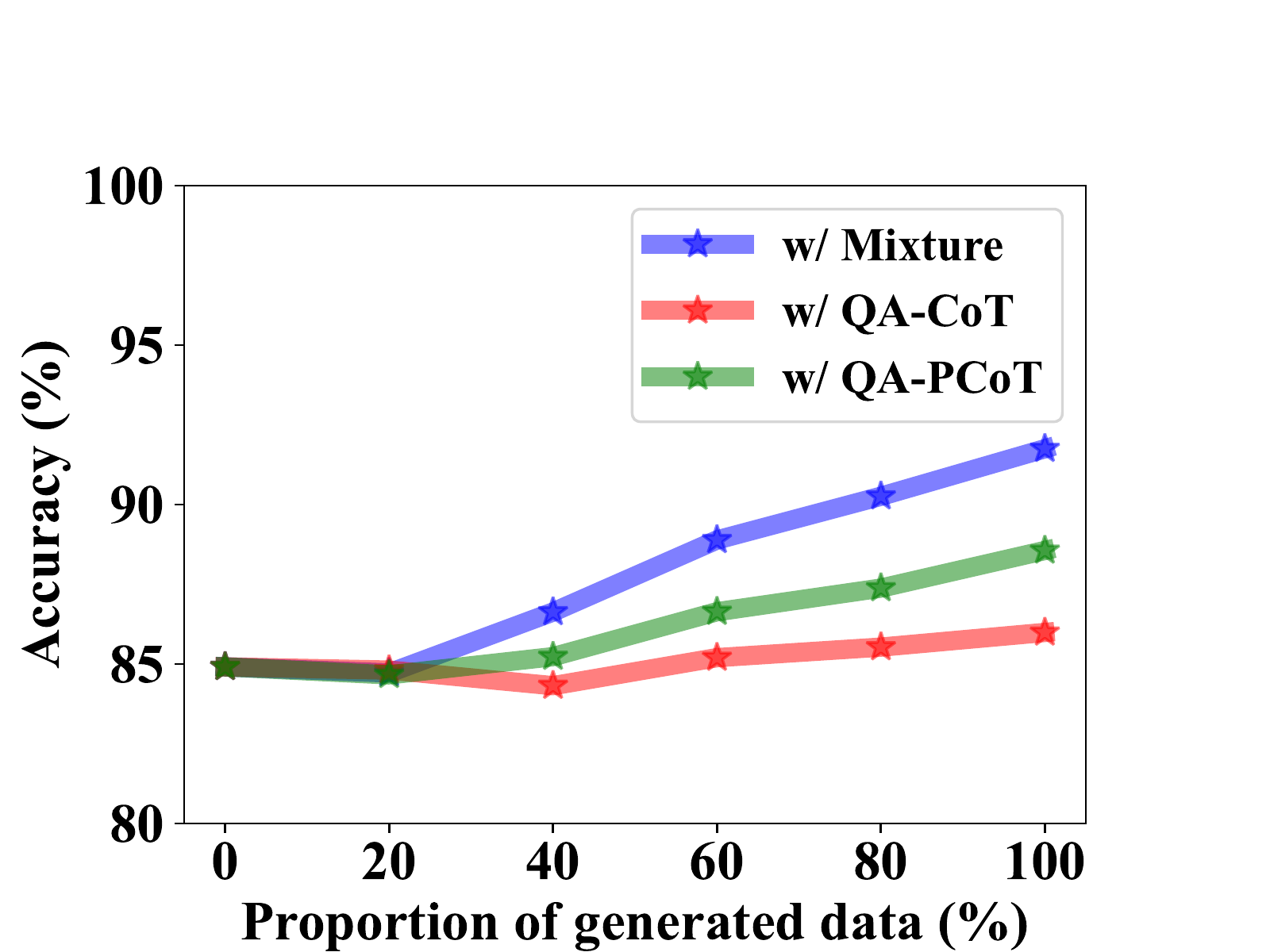}
    \caption{}
    \label{fig:proportion}
    \end{subfigure}
    \begin{subfigure}{0.23\textwidth}
    \centering
    \includegraphics[width=1.0\linewidth]{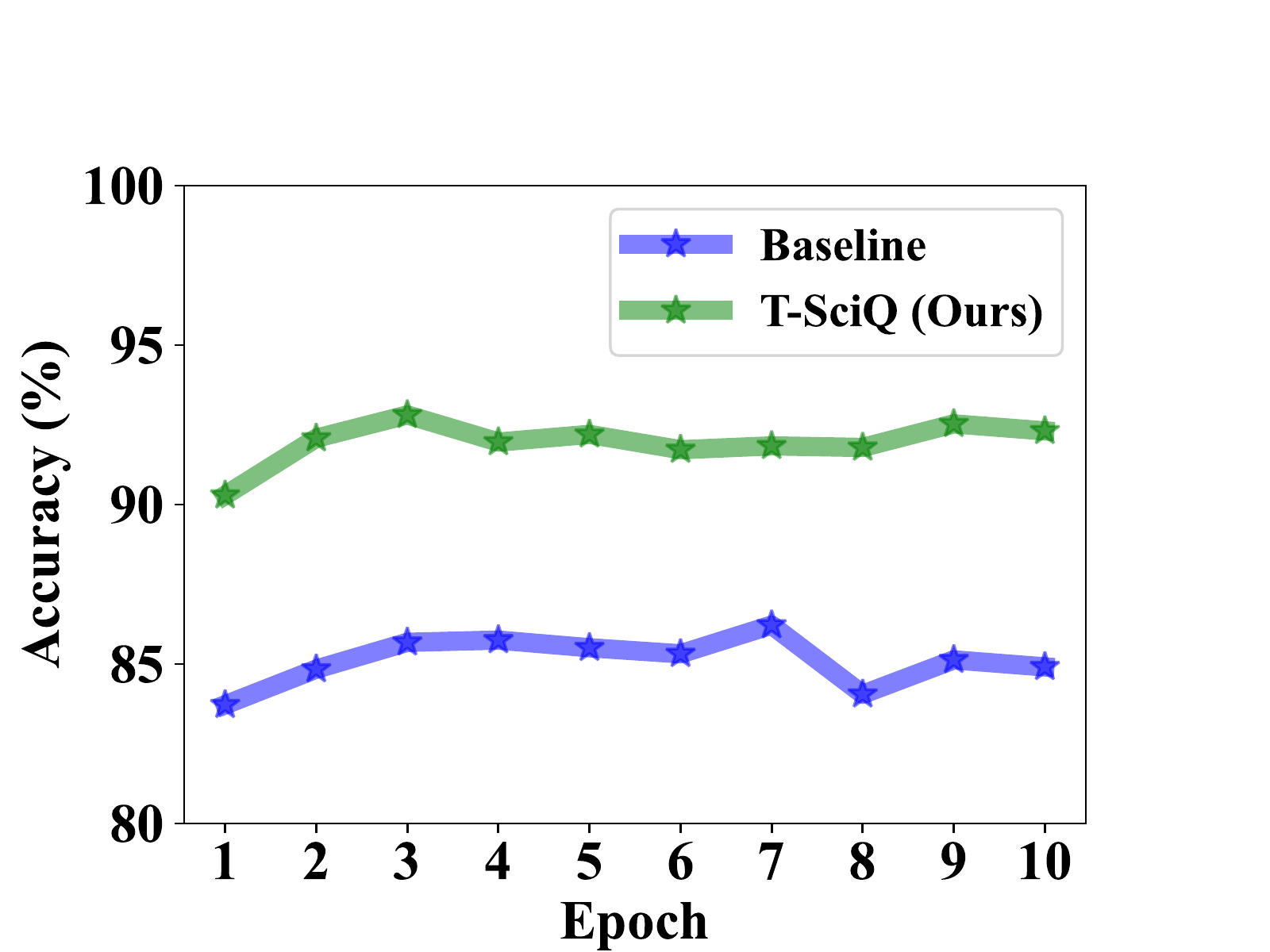}
    \caption{}
    \label{fig:epoch}
    \end{subfigure}
    \vspace{-10pt}
    \caption{
        Further analysis on (a) the effect of Mutimodal-T-SciQ$_\texttt{Base}$ 
 trained with different proportion of generated data and (b) accuracy curve of the baseline Mutimodal-CoT$_\texttt{Base}$ and our Mutimodal-T-SciQ$_\texttt{Base}$  across epochs.
    }
    \vspace{-10pt}
\end{figure}

\begin{figure}[t]
    \centering
    \begin{subfigure}{0.23\textwidth}
    \centering
    \includegraphics[width=1.0\linewidth]{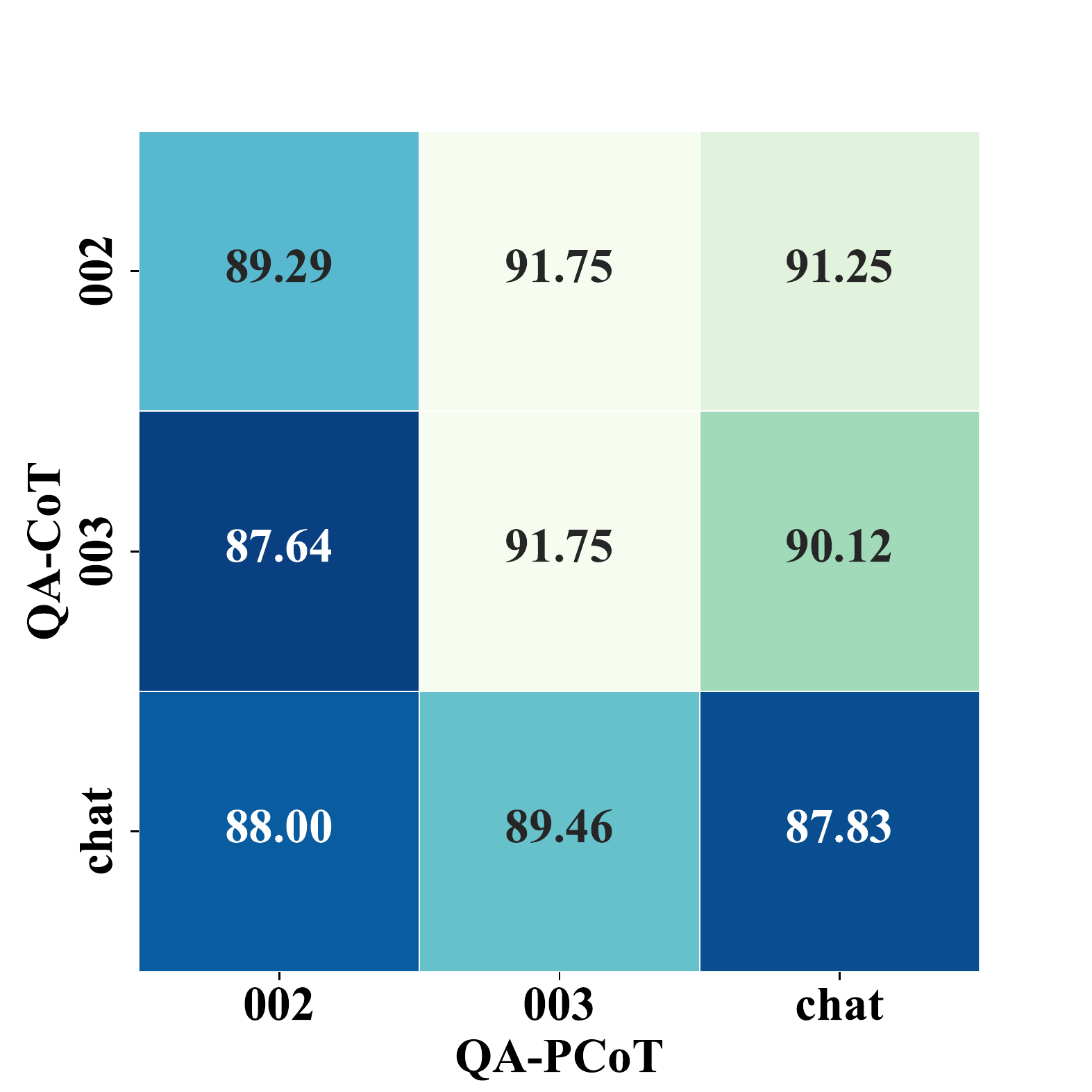}
    \caption{}
    \label{fig:heatmap}
    \end{subfigure}
    \begin{subfigure}{0.23\textwidth}
    \centering
    \includegraphics[width=1.0\linewidth]{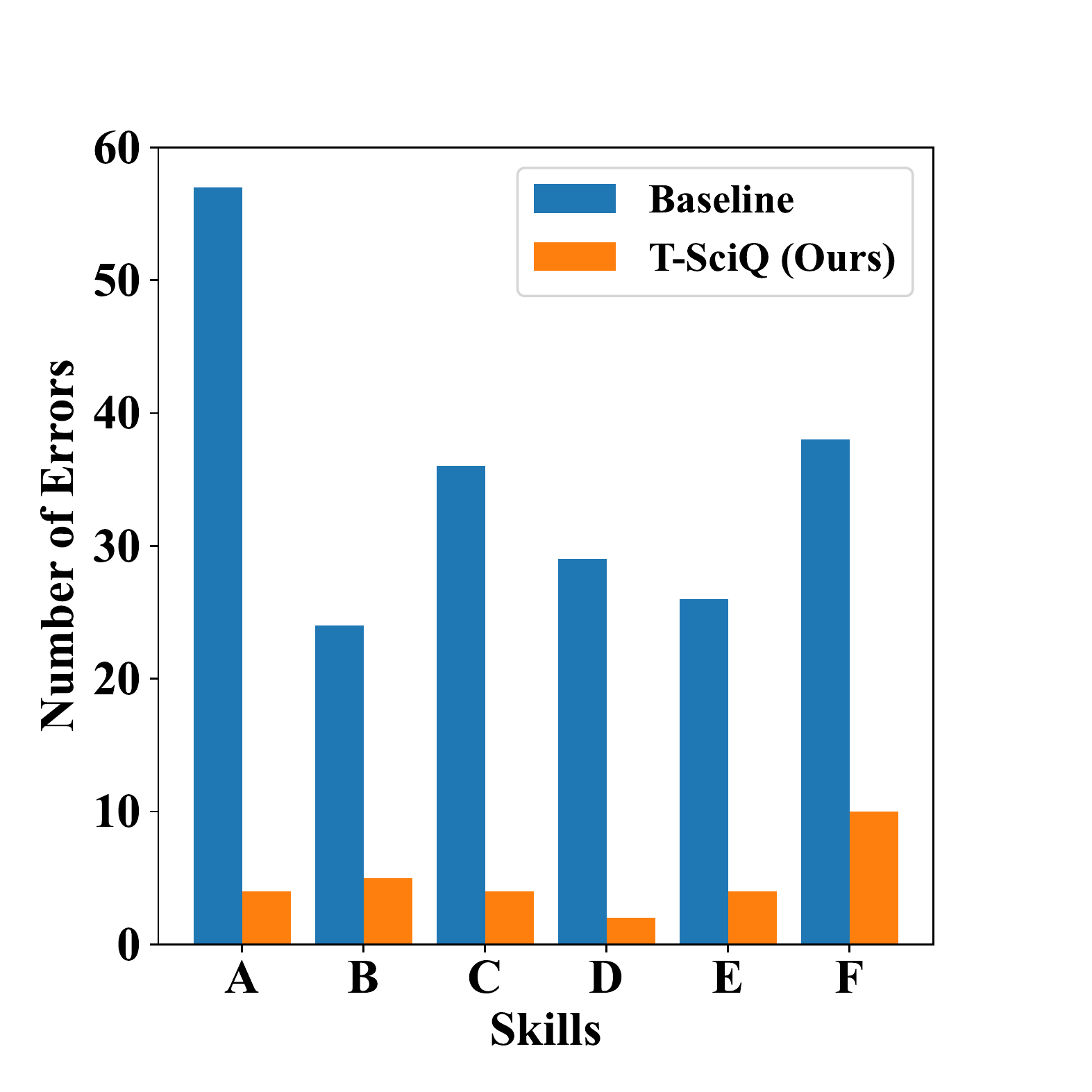}
    \caption{}
    \label{fig:compare}
    \end{subfigure}
    \vspace{-10pt}
    \caption{
        Further analysis on (a) accuracy (\%) of Mutimodal-T-SciQ$_\texttt{Base}$ 
with teaching signals provided by different base LLMs and (b) error analysis of prediction for specific skills.
    }
    \vspace{-10pt}
\end{figure}

\begin{figure*}[!htb]
    \centering
    \begin{subfigure}{0.465\textwidth}
    \centering
    \includegraphics[width=1.0\linewidth]{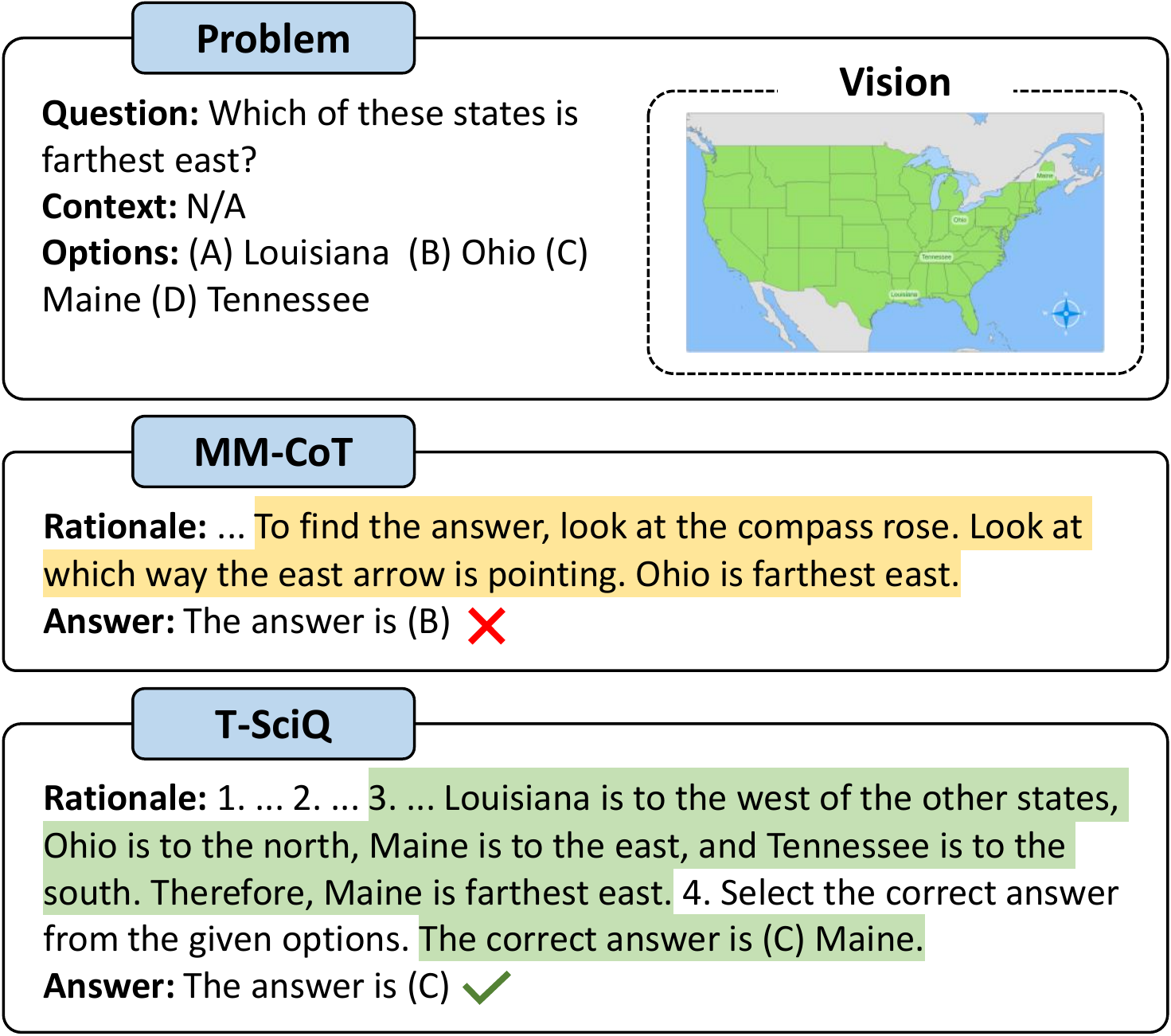}
    \caption{}
    \label{fig:case_study_a}
    \end{subfigure}
    \begin{subfigure}{0.465\textwidth}
    \centering
    \includegraphics[width=1.0\linewidth]{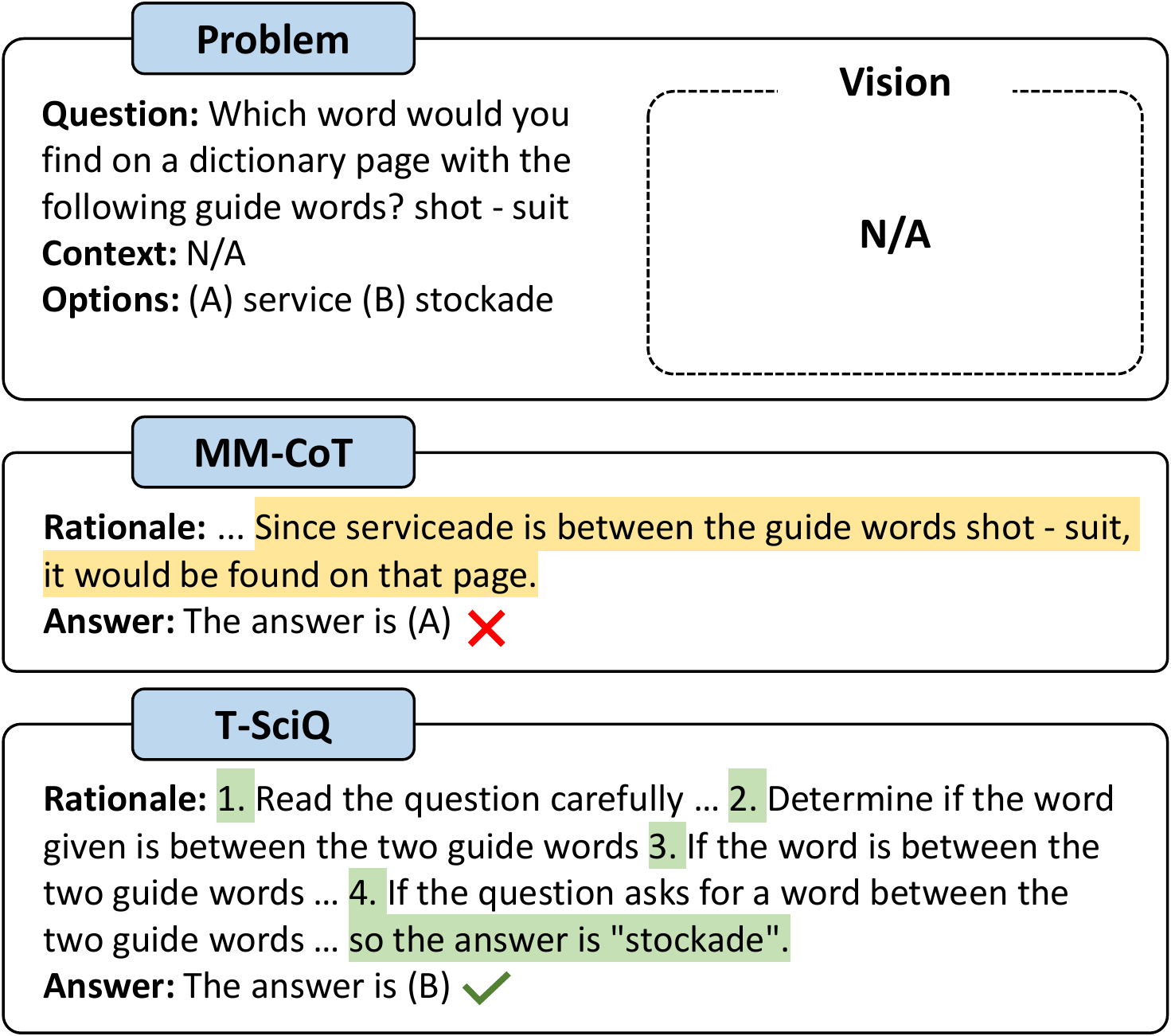}
    \caption{}
    \label{fig:case_study_b}
    \end{subfigure}
    \vspace{-10pt}
    \caption{
        Examples of MM-CoT (baseline) and the model trained with T-SciQ (ours) signals for generating rationales and predicting answers. 
        To solve these examples, commonsense knowledge such as geographic knowledge (a) and multi-step reasoning (b) are required. 
    }
    \label{fig:case_study}
    \vspace{-15pt}
\end{figure*}

\subsection{Further Analysis}
\subsubsection{Effect of Different Signals of T-SciQ.} 
Our approach incorporates two distinct components for teaching signals: QA-CoT and QA-PCoT. We early show that combining these two signals (i.e., Mutimodal-T-SciQ) yields significantly better results than using only human-annotated CoT signals (i.e., Mutimodal-CoT) when teaching student models. In this section, we aim to evaluate the impact of each teaching signals by testing the performance of Mutimodal-T-SciQ$_\texttt{Base}$ and Mutimodal-T-SciQ$_\texttt{Large}$ when either QA-CoT or QA-PCoT signal is removed.
As demonstrated in Table~\ref{tab:ablation_main}, we can observe a significant decrease in answering accuracy when either teaching signal was removed. These findings indicate the effectiveness of both proposed teaching signals. 
This is because 1) student models taught by QA-CoT signals can incorporate a more extensive range of knowledge from the open world rather than solely relying on the knowledge of annotators and 2) student models taught by QA-PCoT signals can decompose complex problems into several simpler sub-problems.

\subsubsection{Impact of visual Features.} 
The choice of visual features can significantly affect the performance of models on ScienceQA. Thus, we conduct an evaluation of three widely-used visual features, which are CLIP \citep{radford2021learning}, DETR \citep{carion2020end}, and ResNet \citep{he2016deep}. Both CLIP and DETR can provide patch-level features, and DETR is designed for object detection. As for ResNet features, we use ResNet-50 to derive visual features.
Table \ref{tab:ablation_vision} shows the results of comparing these three visual features. Our findings suggest that incorporating visual features yields superior performance than relying on language-only baselines. Notably, DETR consistently outperforms the other two features in most cases, and hence, we adopt it as the default visual feature in our main experiments.

\subsubsection{Proportion of Generated Data in Training Data.} 
To further compare the T-SciQ signals produced by LLMs and the annotated CoT signals, we experiment with manipulating the proportion of these two signals within the training data. We vary the proportion of T-SciQ signals from 0\% to 100\%. As demonstrated in Figure~\ref{fig:proportion}, the increasing proportion of training data with T-SciQ signals increases performance. 

\subsubsection{Performance Change with Epoch.} 
Figure \ref{fig:epoch} shows the performance trends of the baseline Mutimodal-CoT$_\texttt{Base}$ and our proposed Mutimodal-T-SciQ$_\texttt{Base}$ across different training epochs. Notably, our method consistently outperforms the baseline across all epochs. We adopt a two-stage training approach similar to the baseline Mutimodal-CoT$_\texttt{Base}$, where we first train the explanation generation module and then train the answer prediction. Hence, like the baseline, our method exhibits relatively higher accuracy at the initial training stages.

\subsubsection{Effect of Teaching Signals Provided by Different Base LLMs.} 
We use the GPT-3.5 model by default, specifically the text-davinci-003 version, to generate teaching signals in the main experiment. However, other powerful LLMs can also provide useful signals, such as the earlier version of GPT-3.5, text-davinci-002, and the recently popular ChatGPT model. This study explores the effectiveness of a mixture of QA-CoT signals from text-davinci-002, text-davinci-003, or ChatGPT, and QA-PCoT signals from the above API-based models. We conduct this experiment using the Multimodal-T-SciQ$_\texttt{Base}$. 
Figure~\ref{fig:heatmap} shows the comparison of the performance of nine different mixture strategies. Our results show that even the worst strategy, which involves a mixture of QA-CoT signal from text-davinci-003 and QA-PCoT signal from text-davince-002, outperforms annotated CoT signal by a significant margin. It indicates that regardless of the mixture strategy used, LLMs can provide signals with more useful knowledge from the open world. 

\subsubsection{Error Analysis.} 
To better understand the model's behavior trained using our proposed T-SciQ signals, we analyze six selected skills shown in Figure~\ref{fig:compare}. It shows the error analysis of prediction for six specific skills  (A-F), i.e., ``\textit{Using guide words}'', ``\textit{Comparing properties of objects}'', ``\textit{Reading a map: cardinal directions}'', ``\textit{Identifying oceans and continents}'', ``\textit{How is temperature related to thermal energy?}'', and ``\textit{Identifying the Thirteen Colonies}'', respectively.
We can observe that training with T-SciQ signals can significantly reduce the number of errors.
Examples of skills such as ``\textit{Identifying oceans and continents}'' require multi-step complex reasoning that T-SciQ teaching signals can teach. On the other hand, examples of skills such as ``\textit{Reading a map: cardinal directions}'' require common sense and factual knowledge from the open world, which T-SciQ signals can also provide.

\subsubsection{Case Study.} 
The case study compares T-SciQ and Multimodal-CoT on the ScienceQA benchmark (Figure~\ref{fig:case_study}). Figure~\ref{fig:case_study_a} shows cases needing geographic knowledge. Human-annotated CoT may lack open-world information, while T-SciQ includes it. Figure~\ref{fig:case_study_b} shows a multi-step reasoning case without image input. Multimodal-CoT errors while our model decomposes and answers correctly. These highlight that T-SciQ is well-suited to handle problems that require open knowledge and decomposition.

\subsection{Comparison on Other NLP Reasoning Datasets.}

\begin{table}[t]
\centering
\small 
\renewcommand\tabcolsep{3pt} 
\caption{Accuracy (\%) on other six reasoning datasets.}
\vspace{-5pt}
\resizebox{0.45\textwidth}{!}{
\begin{tabular}{l|cccccc} 
\toprule
\multirow{2}{*}{Method} & Aqua & Date  & Shuffled  & Coin  &
Common & Strategy \\
& & Unders. & Objects & Flip & SenseQA & QA \\
\midrule
Reason-Teacher & 24.02 & 60.36 & 64.44 & 98.67 & 56.76  & 55.02 \\
T-SciQ & 74.80 & 89.29 & 70.28 & 98.67 & 70.76 & 76.74 \\
\bottomrule
\end{tabular}
}
\vspace{-15pt}
\label{tab:reasonging_teacher}
\end{table}

To verify the versatility of our teaching approach,  we additionally assess our approach on six reasoning tasks, following Reason-Teacher~\cite{ho2022large_teachers}: arithmetic (Aqua~\cite{ling2017program}), symbolic (Coin Flip~\cite{wei2022chain}), commonsense (CommonSenseQA~\cite{talmor2018commonsenseqa}, StrategyQA~\cite{geva2021did}) reasoning, and logic (Date Understanding, Tracking Shuffled Objects)~\cite{geva2021did}.
In Table~\ref{tab:reasonging_teacher}, we compare T-SciQ to diverse reasoning teaching signals introduced by Reason-Teacher. The results show that our T-SciQ surpasses Reason-Teacher by a large margin in 5 out of 6 datasets. It performs equally well in the remaining dataset, Coin Flip. These results indicate that higher-quality teaching signals of planning and reasoning can lead to a remarkable improvement in small student models across different scenarios.

\section{Conclusion}
This paper introduces a new approach named T-SciQ that utilizes large language models' chain-of-thought (CoT) reasoning capabilities to teach small multimodal models for complex science question answering tasks. Our zero-shot prompting method generates QA-CoT samples as teaching data. We also present a 3-step zero-shot prompting approach using plan-based CoT for highly complex problems. Furthermore, our data mixture strategy combines CoT and plan-based CoT to create a new T-SciQ teaching dataset. Empirical evaluation on ScienceQA shows significant improvement over previous state-of-the-art baselines. Our method overcomes the limitations of human-annotated CoT, providing a promising approach for complex science question answering. Future work includes exploring extensive LLMs and parameter-efficient fine-tuning with LLM teachers.

\bibliography{aaai24}

\begin{thebibliography}{56}
\providecommand{\natexlab}[1]{#1}

\bibitem[{Anderson et~al.(2018)Anderson, He, Buehler, Teney, Johnson, Gould,
  and Zhang}]{anderson2018bottom}
Anderson, P.; He, X.; Buehler, C.; Teney, D.; Johnson, M.; Gould, S.; and
  Zhang, L. 2018.
\newblock Bottom-up and top-down attention for image captioning and visual
  question answering.
\newblock In \emph{Proceedings of the IEEE conference on computer vision and
  pattern recognition}, 6077--6086.

\bibitem[{Brown et~al.(2020)Brown, Mann, Ryder, Subbiah, Kaplan, Dhariwal,
  Neelakantan, Shyam, Sastry, Askell et~al.}]{brown2020language}
Brown, T.; Mann, B.; Ryder, N.; Subbiah, M.; Kaplan, J.~D.; Dhariwal, P.;
  Neelakantan, A.; Shyam, P.; Sastry, G.; Askell, A.; et~al. 2020.
\newblock Language models are few-shot learners.
\newblock \emph{Advances in neural information processing systems}, 33:
  1877--1901.

\bibitem[{Carion et~al.(2020)Carion, Massa, Synnaeve, Usunier, Kirillov, and
  Zagoruyko}]{carion2020end}
Carion, N.; Massa, F.; Synnaeve, G.; Usunier, N.; Kirillov, A.; and Zagoruyko,
  S. 2020.
\newblock End-to-End Object Detection with Transformers.
\newblock In \emph{Computer Vision--ECCV 2020: 16th European Conference,
  Glasgow, UK, August 23--28, 2020, Proceedings, Part I}, 213--229.

\bibitem[{Chen et~al.(2020)Chen, Kornblith, Swersky, Norouzi, and
  Hinton}]{chen2020big}
Chen, T.; Kornblith, S.; Swersky, K.; Norouzi, M.; and Hinton, G.~E. 2020.
\newblock Big self-supervised models are strong semi-supervised learners.
\newblock \emph{Advances in neural information processing systems}, 33:
  22243--22255.

\bibitem[{Chen et~al.(2022)Chen, Ma, Wang, and Cohen}]{chen2022program}
Chen, W.; Ma, X.; Wang, X.; and Cohen, W.~W. 2022.
\newblock Program of thoughts prompting: Disentangling computation from
  reasoning for numerical reasoning tasks.
\newblock \emph{arXiv preprint arXiv:2211.12588}.

\bibitem[{Cobbe et~al.(2021)Cobbe, Kosaraju, Bavarian, Chen, Jun, Kaiser,
  Plappert, Tworek, Hilton, Nakano et~al.}]{cobbe2021training}
Cobbe, K.; Kosaraju, V.; Bavarian, M.; Chen, M.; Jun, H.; Kaiser, L.; Plappert,
  M.; Tworek, J.; Hilton, J.; Nakano, R.; et~al. 2021.
\newblock Training verifiers to solve math word problems.
\newblock \emph{arXiv preprint arXiv:2110.14168}.

\bibitem[{Dalvi et~al.(2021)Dalvi, Jansen, Tafjord, Xie, Smith, Pipatanangkura,
  and Clark}]{dalvi2021explaining}
Dalvi, B.; Jansen, P.; Tafjord, O.; Xie, Z.; Smith, H.; Pipatanangkura, L.; and
  Clark, P. 2021.
\newblock Explaining answers with entailment trees.
\newblock \emph{Proceedings of the 2021 Conference on Empirical Methods in
  Natural Language Processing (EMNLP)}.

\bibitem[{Fu et~al.(2023)Fu, Peng, Ou, Sabharwal, and
  Khot}]{fu2023specializing}
Fu, Y.; Peng, H.; Ou, L.; Sabharwal, A.; and Khot, T. 2023.
\newblock Specializing Smaller Language Models towards Multi-Step Reasoning.
\newblock \emph{arXiv preprint arXiv:2301.12726}.

\bibitem[{Fu et~al.(2022)Fu, Peng, Sabharwal, Clark, and
  Khot}]{fu2022complexity}
Fu, Y.; Peng, H.; Sabharwal, A.; Clark, P.; and Khot, T. 2022.
\newblock Complexity-based prompting for multi-step reasoning.
\newblock \emph{arXiv preprint arXiv:2210.00720}.

\bibitem[{Gao et~al.(2019)Gao, Jiang, You, Lu, Hoi, Wang, and
  Li}]{gao2019dynamic}
Gao, P.; Jiang, Z.; You, H.; Lu, P.; Hoi, S.~C.; Wang, X.; and Li, H. 2019.
\newblock Dynamic fusion with intra-and inter-modality attention flow for
  visual question answering.
\newblock In \emph{Proceedings of the IEEE/CVF conference on computer vision
  and pattern recognition}, 6639--6648.

\bibitem[{Geva et~al.(2021)Geva, Khashabi, Segal, Khot, Roth, and
  Berant}]{geva2021did}
Geva, M.; Khashabi, D.; Segal, E.; Khot, T.; Roth, D.; and Berant, J. 2021.
\newblock Did aristotle use a laptop? a question answering benchmark with
  implicit reasoning strategies.
\newblock \emph{Transactions of the Association for Computational Linguistics},
  9: 346--361.

\bibitem[{He et~al.(2023)He, Wang, Hu, Liu, Liu, Xu, and Shen}]{he2023icl}
He, J.; Wang, L.; Hu, Y.; Liu, N.; Liu, H.; Xu, X.; and Shen, H.~T. 2023.
\newblock ICL-D3IE: In-context learning with diverse demonstrations updating
  for document information extraction.
\newblock \emph{arXiv preprint arXiv:2303.05063}.

\bibitem[{He et~al.(2016)He, Zhang, Ren, and Sun}]{he2016deep}
He, K.; Zhang, X.; Ren, S.; and Sun, J. 2016.
\newblock Deep Residual Learning for Image Recognition.
\newblock In \emph{2016 {IEEE} Conference on Computer Vision and Pattern
  Recognition, {CVPR} 2016, Las Vegas, NV, USA, June 27-30, 2016}, 770--778.
  {IEEE} Computer Society.

\bibitem[{Ho, Schmid, and Yun(2022)}]{ho2022large_teachers}
Ho, N.; Schmid, L.; and Yun, S.-Y. 2022.
\newblock Large Language Models Are Reasoning Teachers.
\newblock \emph{arXiv preprint arXiv:2212.10071}.

\bibitem[{Hsieh et~al.(2023)Hsieh, Li, Yeh, Nakhost, Fujii, Ratner, Krishna,
  Lee, and Pfister}]{hsieh2023distilling}
Hsieh, C.-Y.; Li, C.-L.; Yeh, C.-K.; Nakhost, H.; Fujii, Y.; Ratner, A.;
  Krishna, R.; Lee, C.-Y.; and Pfister, T. 2023.
\newblock Distilling step-by-step! outperforming larger language models with
  less training data and smaller model sizes.
\newblock \emph{arXiv preprint arXiv:2305.02301}.

\bibitem[{Hu et~al.(2023)Hu, Lan, Wang, Xu, Lim, Lee, Bing, and
  Poria}]{hu2023llm}
Hu, Z.; Lan, Y.; Wang, L.; Xu, W.; Lim, E.-P.; Lee, R. K.-W.; Bing, L.; and
  Poria, S. 2023.
\newblock LLM-Adapters: An Adapter Family for Parameter-Efficient Fine-Tuning
  of Large Language Models.
\newblock \emph{arXiv preprint arXiv:2304.01933}.

\bibitem[{Huang et~al.(2022)Huang, Gu, Hou, Wu, Wang, Yu, and
  Han}]{huang2022large}
Huang, J.; Gu, S.~S.; Hou, L.; Wu, Y.; Wang, X.; Yu, H.; and Han, J. 2022.
\newblock Large language models can self-improve.
\newblock \emph{arXiv preprint arXiv:2210.11610}.

\bibitem[{Jansen et~al.(2018)Jansen, Wainwright, Marmorstein, and
  Morrison}]{jansen2018worldtree}
Jansen, P.~A.; Wainwright, E.; Marmorstein, S.; and Morrison, C.~T. 2018.
\newblock Worldtree: A corpus of explanation graphs for elementary science
  questions supporting multi-hop inference.
\newblock \emph{arXiv preprint arXiv:1802.03052}.

\bibitem[{Kembhavi et~al.(2017)Kembhavi, Seo, Schwenk, Choi, Farhadi, and
  Hajishirzi}]{kembhavi2017you}
Kembhavi, A.; Seo, M.; Schwenk, D.; Choi, J.; Farhadi, A.; and Hajishirzi, H.
  2017.
\newblock Are you smarter than a sixth grader? textbook question answering for
  multimodal machine comprehension.
\newblock In \emph{Proceedings of the IEEE Conference on Computer Vision and
  Pattern Recognition (CVPR)}, 4999--5007.

\bibitem[{Khashabi et~al.(2020)Khashabi, Min, Khot, Sabharwal, Tafjord, Clark,
  and Hajishirzi}]{khashabi2020unifiedqa}
Khashabi, D.; Min, S.; Khot, T.; Sabharwal, A.; Tafjord, O.; Clark, P.; and
  Hajishirzi, H. 2020.
\newblock Unifiedqa: Crossing format boundaries with a single qa system.
\newblock \emph{arXiv preprint arXiv:2005.00700}.

\bibitem[{Khot et~al.(2022)Khot, Trivedi, Finlayson, Fu, Richardson, Clark, and
  Sabharwal}]{khot2022decomposed}
Khot, T.; Trivedi, H.; Finlayson, M.; Fu, Y.; Richardson, K.; Clark, P.; and
  Sabharwal, A. 2022.
\newblock Decomposed prompting: A modular approach for solving complex tasks.
\newblock \emph{arXiv preprint arXiv:2210.02406}.

\bibitem[{Kim, Jun, and Zhang(2018)}]{kim2018bilinear}
Kim, J.-H.; Jun, J.; and Zhang, B.-T. 2018.
\newblock Bilinear attention networks.
\newblock \emph{Advances in neural information processing systems}, 31.

\bibitem[{Kim, Son, and Kim(2021)}]{kim2021vilt}
Kim, W.; Son, B.; and Kim, I. 2021.
\newblock Vilt: Vision-and-language transformer without convolution or region
  supervision.
\newblock In \emph{International Conference on Machine Learning}, 5583--5594.
  PMLR.

\bibitem[{Kojima et~al.(2022)Kojima, Gu, Reid, Matsuo, and
  Iwasawa}]{kojima2022large}
Kojima, T.; Gu, S.~S.; Reid, M.; Matsuo, Y.; and Iwasawa, Y. 2022.
\newblock Large language models are zero-shot reasoners.
\newblock \emph{arXiv preprint arXiv:2205.11916}.

\bibitem[{Li et~al.(2022{\natexlab{a}})Li, Lv, Zhou, Zhou, Xiao, Ma, and
  Zhu}]{li2022vision}
Li, B.; Lv, C.; Zhou, Z.; Zhou, T.; Xiao, T.; Ma, A.; and Zhu, J.
  2022{\natexlab{a}}.
\newblock On Vision Features in Multimodal Machine Translation.
\newblock In \emph{Proceedings of the 60th Annual Meeting of the Association
  for Computational Linguistics (Volume 1: Long Papers)}, 6327--6337.

\bibitem[{Li et~al.(2019)Li, Yatskar, Yin, Hsieh, and Chang}]{li2019visualbert}
Li, L.~H.; Yatskar, M.; Yin, D.; Hsieh, C.-J.; and Chang, K.-W. 2019.
\newblock Visualbert: A simple and performant baseline for vision and language.
\newblock \emph{arXiv preprint arXiv:1908.03557}.

\bibitem[{Li et~al.(2022{\natexlab{b}})Li, Lin, Zhang, Fu, Chen, Lou, and
  Chen}]{li2022advance}
Li, Y.; Lin, Z.; Zhang, S.; Fu, Q.; Chen, B.; Lou, J.-G.; and Chen, W.
  2022{\natexlab{b}}.
\newblock On the advance of making language models better reasoners.
\newblock \emph{arXiv preprint arXiv:2206.02336}.

\bibitem[{Ling et~al.(2017)Ling, Yogatama, Dyer, and Blunsom}]{ling2017program}
Ling, W.; Yogatama, D.; Dyer, C.; and Blunsom, P. 2017.
\newblock Program induction by rationale generation: Learning to solve and
  explain algebraic word problems.
\newblock \emph{arXiv preprint arXiv:1705.04146}.

\bibitem[{Liu et~al.(2023)Liu, Li, Wu, and Lee}]{liu2023visual_instruction}
Liu, H.; Li, C.; Wu, Q.; and Lee, Y.~J. 2023.
\newblock Visual instruction tuning.
\newblock \emph{arXiv preprint arXiv:2304.08485}.

\bibitem[{Lu et~al.(2022{\natexlab{a}})Lu, Mishra, Xia, Qiu, Chang, Zhu,
  Tafjord, Clark, and Kalyan}]{lu2022learn}
Lu, P.; Mishra, S.; Xia, T.; Qiu, L.; Chang, K.-W.; Zhu, S.-C.; Tafjord, O.;
  Clark, P.; and Kalyan, A. 2022{\natexlab{a}}.
\newblock Learn to explain: Multimodal reasoning via thought chains for science
  question answering.
\newblock \emph{Advances in Neural Information Processing Systems}, 35:
  2507--2521.

\bibitem[{Lu et~al.(2023)Lu, Peng, Cheng, Galley, Chang, Wu, Zhu, and
  Gao}]{lu2023chameleon}
Lu, P.; Peng, B.; Cheng, H.; Galley, M.; Chang, K.-W.; Wu, Y.~N.; Zhu, S.-C.;
  and Gao, J. 2023.
\newblock Chameleon: Plug-and-Play Compositional Reasoning with Large Language
  Models.
\newblock \emph{arXiv preprint arXiv:2304.09842}.

\bibitem[{Lu et~al.(2022{\natexlab{b}})Lu, Qiu, Chang, Wu, Zhu, Rajpurohit,
  Clark, and Kalyan}]{lu2022dynamic}
Lu, P.; Qiu, L.; Chang, K.-W.; Wu, Y.~N.; Zhu, S.-C.; Rajpurohit, T.; Clark,
  P.; and Kalyan, A. 2022{\natexlab{b}}.
\newblock Dynamic prompt learning via policy gradient for semi-structured
  mathematical reasoning.
\newblock \emph{arXiv preprint arXiv:2209.14610}.

\bibitem[{Lu et~al.(2021)Lu, Qiu, Chen, Xia, Zhao, Zhang, Yu, Liang, and
  Zhu}]{lu2021iconqa}
Lu, P.; Qiu, L.; Chen, J.; Xia, T.; Zhao, Y.; Zhang, W.; Yu, Z.; Liang, X.; and
  Zhu, S.-C. 2021.
\newblock Iconqa: A new benchmark for abstract diagram understanding and visual
  language reasoning.
\newblock \emph{arXiv preprint arXiv:2110.13214}.

\bibitem[{Magister et~al.(2022)Magister, Mallinson, Adamek, Malmi, and
  Severyn}]{magister2022teaching}
Magister, L.~C.; Mallinson, J.; Adamek, J.; Malmi, E.; and Severyn, A. 2022.
\newblock Teaching small language models to reason.
\newblock \emph{arXiv preprint arXiv:2212.08410}.

\bibitem[{Nye et~al.(2021)Nye, Andreassen, Gur-Ari, Michalewski, Austin,
  Bieber, Dohan, Lewkowycz, Bosma, Luan et~al.}]{nye2021show}
Nye, M.; Andreassen, A.~J.; Gur-Ari, G.; Michalewski, H.; Austin, J.; Bieber,
  D.; Dohan, D.; Lewkowycz, A.; Bosma, M.; Luan, D.; et~al. 2021.
\newblock Show your work: Scratchpads for intermediate computation with
  language models.
\newblock \emph{arXiv preprint arXiv:2112.00114}.

\bibitem[{OpenAI(2022)}]{openai-chatgpt-2022}
OpenAI. 2022.
\newblock Introducing chatgpt.
\newblock \url{https://openai.com/blog/chatgpt}.

\bibitem[{OpenAI(2023)}]{openai-gpt4-2023}
OpenAI. 2023.
\newblock {GPT-4} Technical Report.
\newblock \emph{CoRR}, abs/2303.08774.

\bibitem[{Radford et~al.(2021)Radford, Kim, Hallacy, Ramesh, Goh, Agarwal,
  Sastry, Askell, Mishkin, Clark et~al.}]{radford2021learning}
Radford, A.; Kim, J.~W.; Hallacy, C.; Ramesh, A.; Goh, G.; Agarwal, S.; Sastry,
  G.; Askell, A.; Mishkin, P.; Clark, J.; et~al. 2021.
\newblock Learning transferable visual models from natural language
  supervision.
\newblock In \emph{International Conference on Machine Learning}, 8748--8763.
  PMLR.

\bibitem[{Rubin, Herzig, and Berant(2021)}]{rubin2021learning}
Rubin, O.; Herzig, J.; and Berant, J. 2021.
\newblock Learning to retrieve prompts for in-context learning.
\newblock \emph{arXiv preprint arXiv:2112.08633}.

\bibitem[{Sampat, Yang, and Baral(2020)}]{sampat2020visuo}
Sampat, S.~K.; Yang, Y.; and Baral, C. 2020.
\newblock Visuo-Lingustic Question Answering (VLQA) Challenge.
\newblock In \emph{Proceedings of the 2020 Conference on Empirical Methods in
  Natural Language Processing: Findings (EMNLP)}, 4606--4616.

\bibitem[{Talmor et~al.(2018)Talmor, Herzig, Lourie, and
  Berant}]{talmor2018commonsenseqa}
Talmor, A.; Herzig, J.; Lourie, N.; and Berant, J. 2018.
\newblock Commonsenseqa: A question answering challenge targeting commonsense
  knowledge.
\newblock \emph{arXiv preprint arXiv:1811.00937}.

\bibitem[{Thoppilan et~al.(2022)Thoppilan, De~Freitas, Hall, Shazeer,
  Kulshreshtha, Cheng, Jin, Bos, Baker, Du et~al.}]{thoppilan2022lamda}
Thoppilan, R.; De~Freitas, D.; Hall, J.; Shazeer, N.; Kulshreshtha, A.; Cheng,
  H.-T.; Jin, A.; Bos, T.; Baker, L.; Du, Y.; et~al. 2022.
\newblock Lamda: Language models for dialog applications.
\newblock \emph{arXiv preprint arXiv:2201.08239}.

\bibitem[{Tian et~al.(2023)Tian, Zhu, Wang, Li, and Lan}]{tian2023r}
Tian, Q.; Zhu, H.; Wang, L.; Li, Y.; and Lan, Y. 2023.
\newblock R$^3$ Prompting: Review, Rephrase and Resolve for Chain-of-Thought
  Reasoning in Large Language Models under Noisy Context.
\newblock \emph{arXiv preprint arXiv:2310.16535}.

\bibitem[{Touvron et~al.(2023)Touvron, Lavril, Izacard, Martinet, Lachaux,
  Lacroix, Rozi{\`e}re, Goyal, Hambro, Azhar et~al.}]{touvron2023llama}
Touvron, H.; Lavril, T.; Izacard, G.; Martinet, X.; Lachaux, M.-A.; Lacroix,
  T.; Rozi{\`e}re, B.; Goyal, N.; Hambro, E.; Azhar, F.; et~al. 2023.
\newblock Llama: Open and efficient foundation language models.
\newblock \emph{arXiv preprint arXiv:2302.13971}.

\bibitem[{Vaswani et~al.(2017)Vaswani, Shazeer, Parmar, Uszkoreit, Jones,
  Gomez, Kaiser, and Polosukhin}]{transformer}
Vaswani, A.; Shazeer, N.; Parmar, N.; Uszkoreit, J.; Jones, L.; Gomez, A.~N.;
  Kaiser, L.; and Polosukhin, I. 2017.
\newblock Attention is All you Need.
\newblock In \emph{Advances in Neural Information Processing Systems 30},
  5998--6008.

\bibitem[{Wang, Deng, and Sun(2022)}]{wang2022iteratively}
Wang, B.; Deng, X.; and Sun, H. 2022.
\newblock Iteratively prompt pre-trained language models for chain of thought.
\newblock In \emph{Proceedings of the 2022 Conference on Empirical Methods in
  Natural Language Processing}, 2714--2730.

\bibitem[{Wang et~al.(2023)Wang, Xu, Lan, Hu, Lan, Lee, and Lim}]{wang2023plan}
Wang, L.; Xu, W.; Lan, Y.; Hu, Z.; Lan, Y.; Lee, R. K.-W.; and Lim, E.-P. 2023.
\newblock Plan-and-solve prompting: Improving zero-shot chain-of-thought
  reasoning by large language models.
\newblock \emph{arXiv preprint arXiv:2305.04091}.

\bibitem[{Wang et~al.(2022{\natexlab{a}})Wang, Wei, Schuurmans, Le, Chi, and
  Zhou}]{wang2022rationale}
Wang, X.; Wei, J.; Schuurmans, D.; Le, Q.; Chi, E.; and Zhou, D.
  2022{\natexlab{a}}.
\newblock Rationale-augmented ensembles in language models.
\newblock \emph{arXiv preprint arXiv:2207.00747}.

\bibitem[{Wang et~al.(2022{\natexlab{b}})Wang, Wei, Schuurmans, Le, Chi, and
  Zhou}]{wang2022self}
Wang, X.; Wei, J.; Schuurmans, D.; Le, Q.; Chi, E.; and Zhou, D.
  2022{\natexlab{b}}.
\newblock Self-consistency improves chain of thought reasoning in language
  models.
\newblock \emph{arXiv preprint arXiv:2203.11171}.

\bibitem[{Wei et~al.(2022{\natexlab{a}})Wei, Wang, Schuurmans, Bosma, Chi, Le,
  and Zhou}]{cot_wei}
Wei, J.; Wang, X.; Schuurmans, D.; Bosma, M.; Chi, E.; Le, Q.; and Zhou, D.
  2022{\natexlab{a}}.
\newblock Chain of Thought Prompting Elicits Reasoning in Large Language
  Models.
\newblock \emph{ArXiv preprint}, abs/2201.11903.

\bibitem[{Wei et~al.(2022{\natexlab{b}})Wei, Wang, Schuurmans, Bosma, Chi, Le,
  and Zhou}]{wei2022chain}
Wei, J.; Wang, X.; Schuurmans, D.; Bosma, M.; Chi, E.; Le, Q.; and Zhou, D.
  2022{\natexlab{b}}.
\newblock Chain of thought prompting elicits reasoning in large language
  models.
\newblock \emph{arXiv preprint arXiv:2201.11903}.

\bibitem[{Yu et~al.(2019)Yu, Yu, Cui, Tao, and Tian}]{yu2019deep}
Yu, Z.; Yu, J.; Cui, Y.; Tao, D.; and Tian, Q. 2019.
\newblock Deep modular co-attention networks for visual question answering.
\newblock In \emph{Proceedings of the IEEE/CVF conference on computer vision
  and pattern recognition}, 6281--6290.

\bibitem[{Zhang et~al.(2023{\natexlab{a}})Zhang, Han, Zhou, Hu, Yan, Lu, Li,
  Gao, and Qiao}]{zhang2023llama_adapter}
Zhang, R.; Han, J.; Zhou, A.; Hu, X.; Yan, S.; Lu, P.; Li, H.; Gao, P.; and
  Qiao, Y. 2023{\natexlab{a}}.
\newblock Llama-adapter: Efficient fine-tuning of language models with
  zero-init attention.
\newblock \emph{arXiv preprint arXiv:2303.16199}.

\bibitem[{Zhang et~al.(2022)Zhang, Zhang, Li, and Smola}]{zhang2022automatic}
Zhang, Z.; Zhang, A.; Li, M.; and Smola, A. 2022.
\newblock Automatic chain of thought prompting in large language models.
\newblock \emph{arXiv preprint arXiv:2210.03493}.

\bibitem[{Zhang et~al.(2023{\natexlab{b}})Zhang, Zhang, Li, Zhao, Karypis, and
  Smola}]{zhang2023multimodal}
Zhang, Z.; Zhang, A.; Li, M.; Zhao, H.; Karypis, G.; and Smola, A.
  2023{\natexlab{b}}.
\newblock Multimodal chain-of-thought reasoning in language models.
\newblock \emph{arXiv preprint arXiv:2302.00923}.

\bibitem[{Zhou et~al.(2022)Zhou, Sch{\"a}rli, Hou, Wei, Scales, Wang,
  Schuurmans, Bousquet, Le, and Chi}]{zhou2022least}
Zhou, D.; Sch{\"a}rli, N.; Hou, L.; Wei, J.; Scales, N.; Wang, X.; Schuurmans,
  D.; Bousquet, O.; Le, Q.; and Chi, E. 2022.
\newblock Least-to-most prompting enables complex reasoning in large language
  models.
\newblock \emph{arXiv preprint arXiv:2205.10625}.

\end{thebibliography}

\newpage
\appendix
\onecolumn

\section{More Detailed Analysis}

In addition to exploring the effects of different visual features, we also tested the performance of different backbones. Table~\ref{tab:backbone} shows the results of UnifiedQA and FLAN-T5 on our generated datasets. When using various backbone networks, the model trained by QA-CoT data is the worst among the three generated data types. However, it is also better than the manually annotated data, which indicates that the manually annotated data has certain limitations, such as redundant information, single style, etc. Furthermore, when using the Mixture dataset of QA-CoT and QA-PCoT, the fact that all four backbones achieved the best performances illustrates the effectiveness and generality of our strategy.

\begin{table}[h]
\centering
\caption{Accuracy (\%) of using different backbones.}
\small
\renewcommand\tabcolsep{3pt} 
\resizebox{0.47\textwidth}{!}{
\begin{tabular}{l|c|c|ccc} 
\toprule
 \multirow{2}{*}{Method} & \multirow{2}{*}{Size} &\multirow{2}{*}{Annotated CoT} & \multicolumn{3}{c}{T-SciQ} \\
 & && QA-CoT & QA-PCoT & Mixture \\
\midrule
UnifiedQA$_\texttt{Base}$ & 223M & 84.91 & 85.99 & 88.56 & 91.75 \\
UnifiedQA$_\texttt{Large}$ & 738M & 91.68 & 93.44 & 93.54 & 96.18 \\
\midrule
FLAN-T5$_\texttt{Base}$ & 248M & 85.85 & 86.87 & 89.04 & 92.33 \\
FLAN-T5$_\texttt{Large}$ & 783M & 93.02 &93.54 & 95.68 & \bf 96.49 \\
\bottomrule
\end{tabular}
}
\label{tab:backbone}
\end{table}

\section{More Cases Analysis}

\begin{figure*}[!htb]
    \centering
    \begin{subfigure}{0.992\textwidth}
    \centering
    \includegraphics[width=1.0\linewidth]{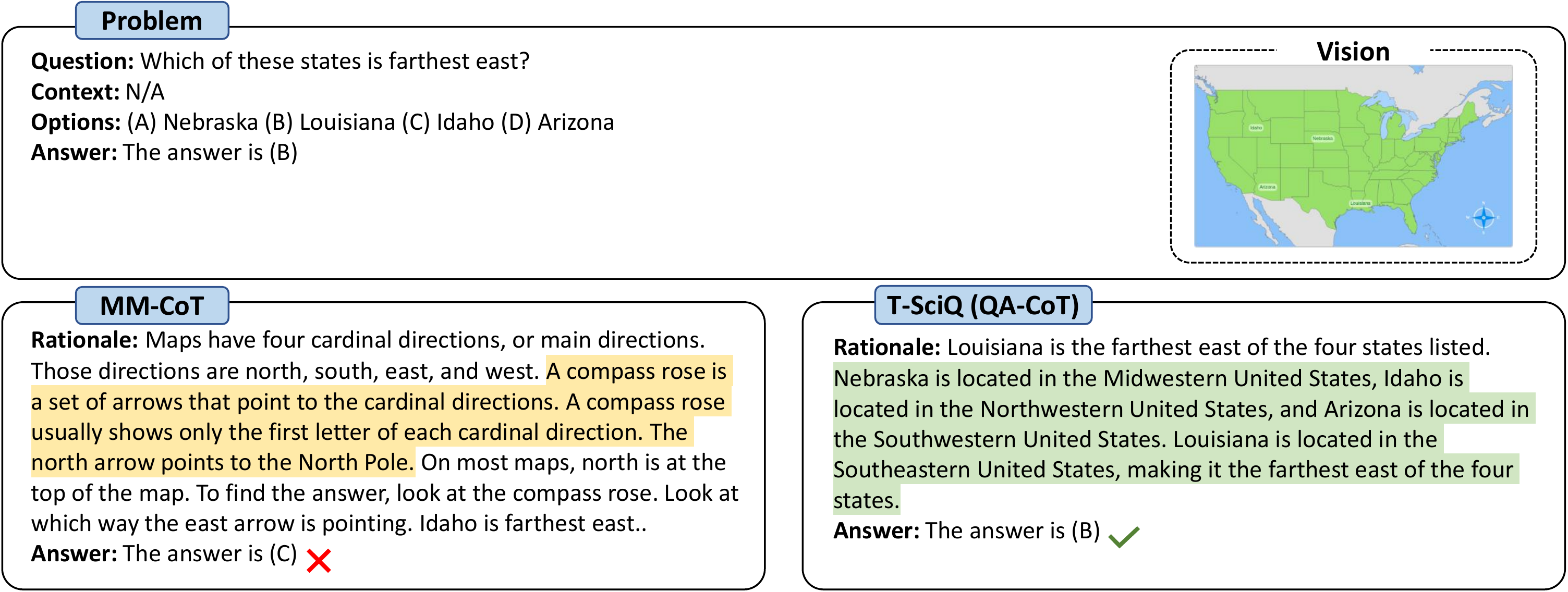}
    \caption{
        There is some redundancy in manually annotating data.
    }
    \label{fig:case_study_supp_qar_baseline1}
    \end{subfigure}
    \\
    \begin{subfigure}{0.992\textwidth}
    \centering
    \includegraphics[width=1.0\linewidth]{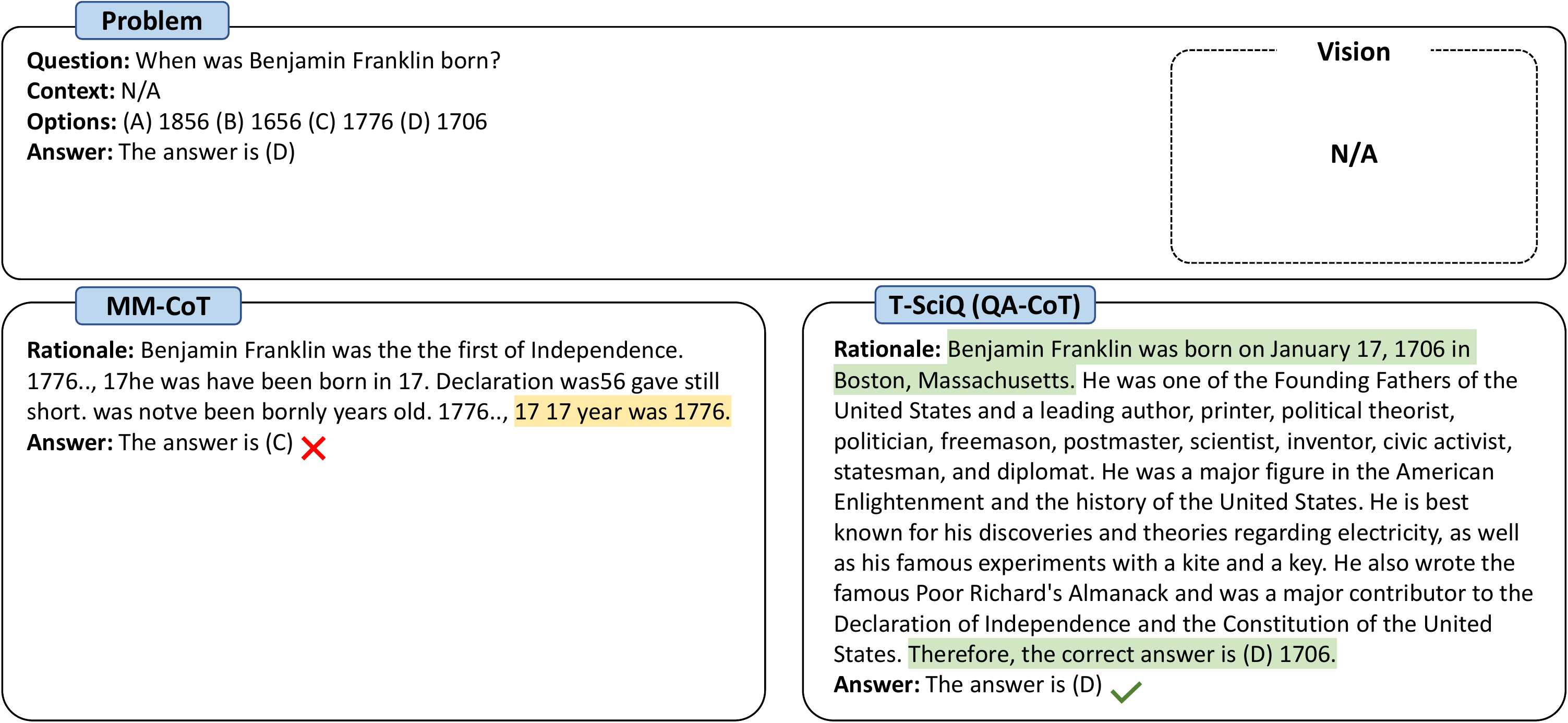}
    \caption{
        The QA-CoT data can be used to solve problems with commonsense knowledge.
    }
    \label{fig:case_study_supp_qar_baseline2}
    \end{subfigure}
    \caption{
        Examples of MM-CoT (baseline) and T-SciQ (QA-CoT) (ours) for generating rationales and predicting answers.
    }
    \label{fig:case_study_supp_qar_baseline}
\end{figure*}

\begin{figure*}[!htb]
    \centering
    \begin{subfigure}{0.992\textwidth}
    \centering
    \includegraphics[width=1.0\linewidth]{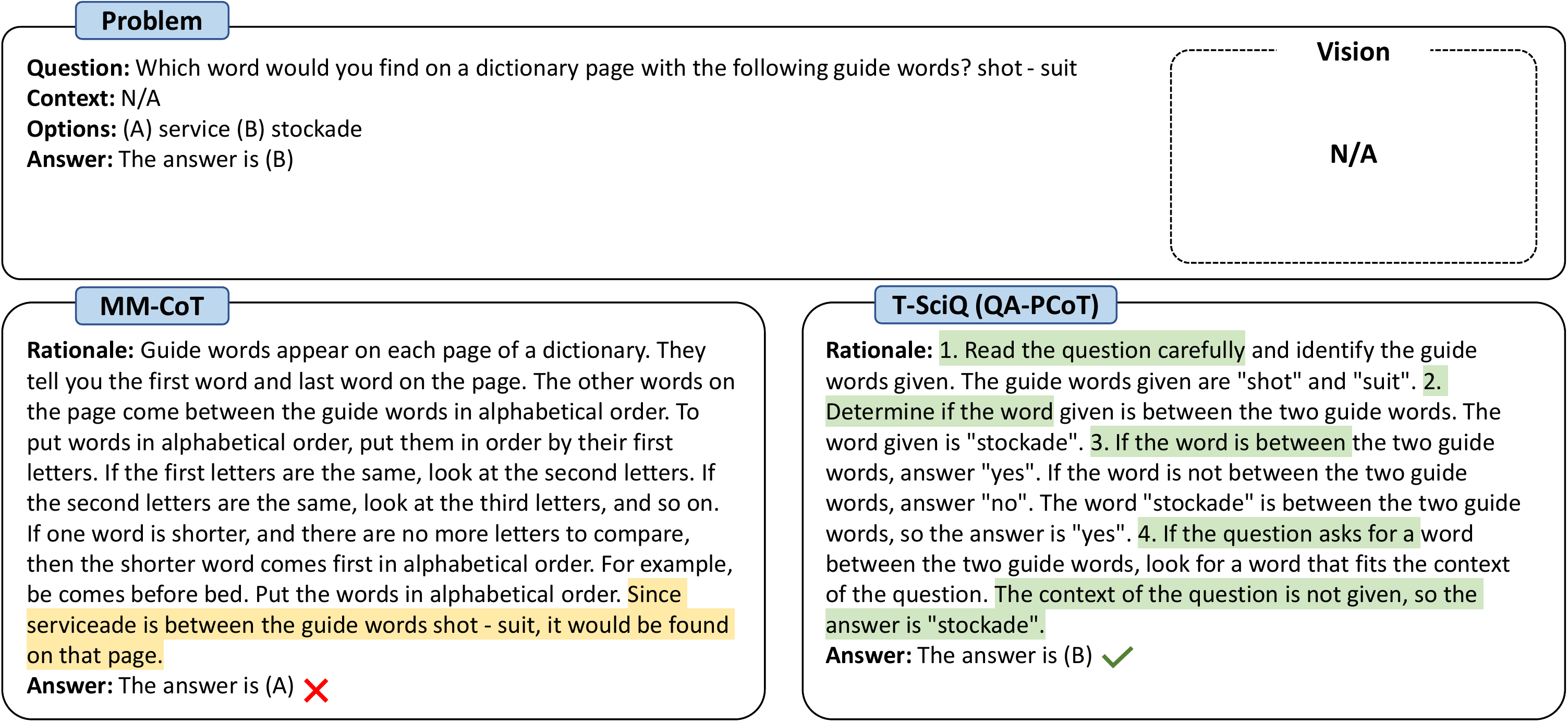}
    \caption{
        T-SciQ (QA-PCoT) can solve problems by executing the plan in multiple steps.
    }
    \label{fig:case_study_supp_ps_baseline1}
    \end{subfigure}
    \\
    \begin{subfigure}{0.992\textwidth}
    \centering
    \includegraphics[width=1.0\linewidth]{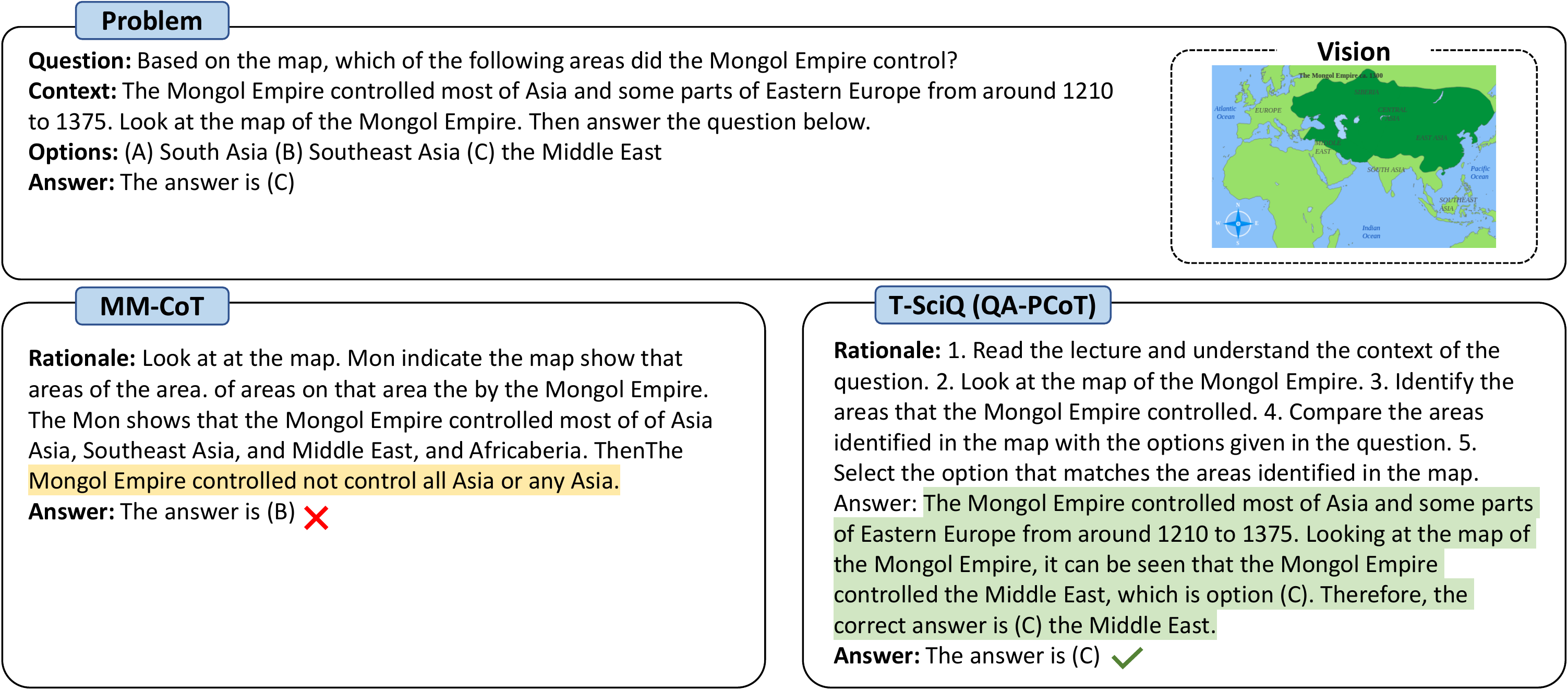}
    \caption{
       The QA-PCoT data can be used to solve problems with commonsense knowledge.
    }
    \label{fig:case_study_supp_ps_baseline2}
    \end{subfigure}
    \caption{
        Examples of MM-CoT (baseline) and T-SciQ (QA-PCoT) (ours) for generating rationales and predicting answers.
    }
    \label{fig:case_study_supp_ps_baseline}
\end{figure*}

\begin{figure*}[!htb]
    \centering
    \begin{subfigure}{0.992\textwidth}
    \centering
    \includegraphics[width=1.0\linewidth]{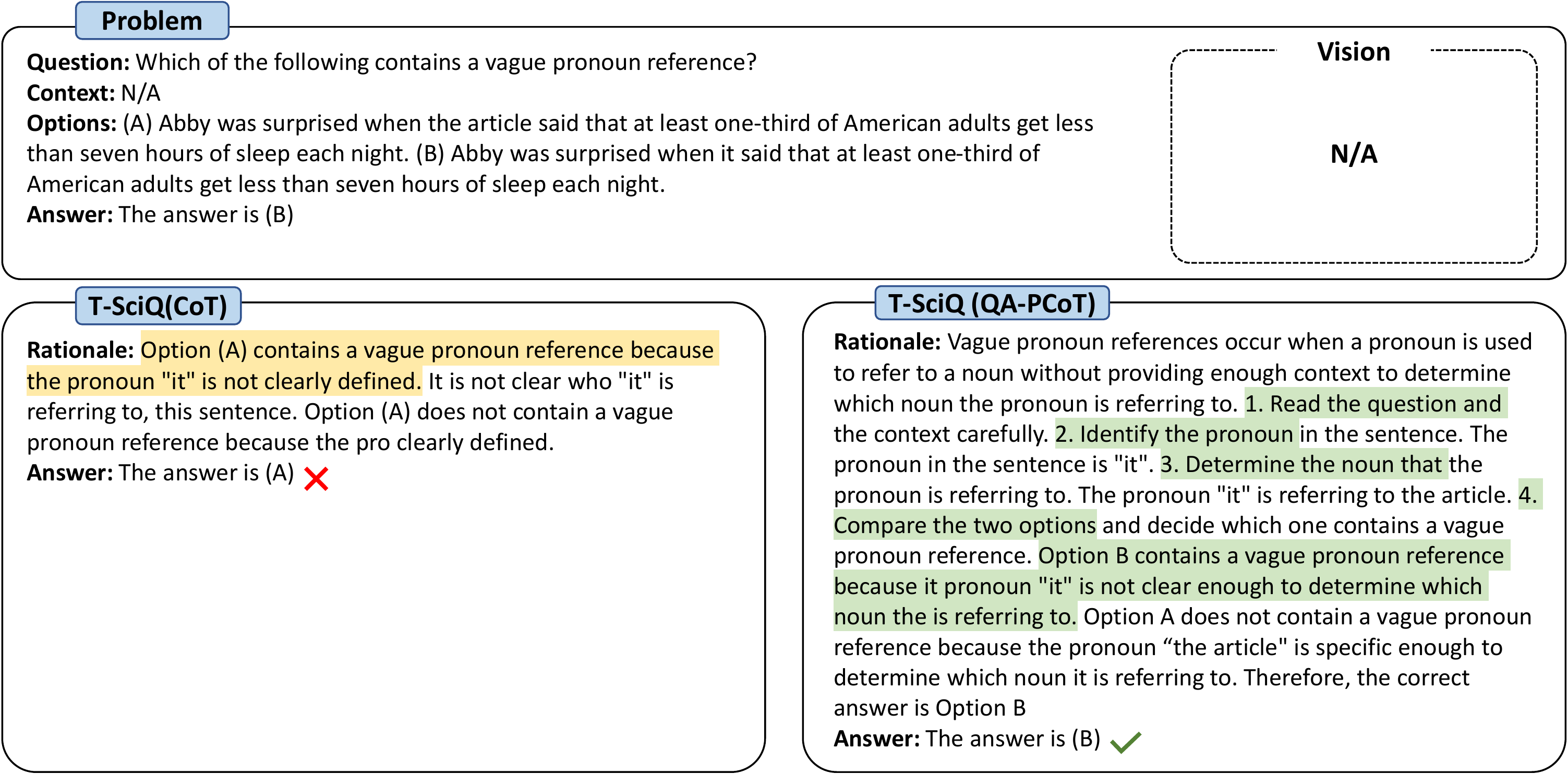}
    \caption{
        T-SciQ (QA-PCoT) can solve problems by executing the plan in multiple steps.
    }
    \label{fig:case_study_supp_ps_qar1}
    \end{subfigure}
    \\
    \begin{subfigure}{0.992\textwidth}
    \centering
    \includegraphics[width=1.0\linewidth]{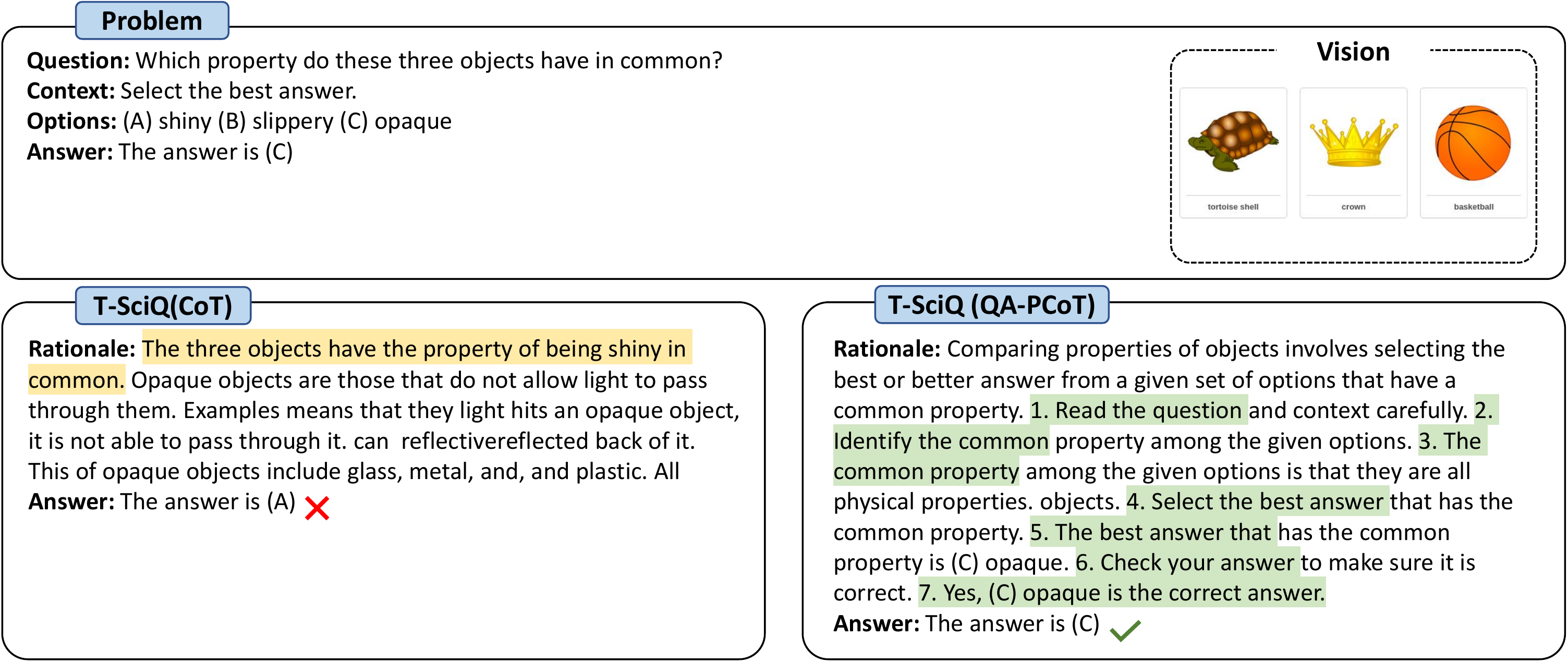}
    \caption{
        T-SciQ (QA-PCoT) can solve problems by executing the plan in multiple steps.
    }
    \label{fig:case_study_supp_ps_qar2}
    \end{subfigure}
    \caption{
        Examples of T-SciQ (QA-CoT) (ours) and T-SciQ (QA-PCoT) (ours) for generating rationales and predicting answers.
    }
    \label{fig:case_study_supp_ps_qar}
\end{figure*}

\begin{figure*}[!htb]
    \centering
    \begin{subfigure}{0.992\textwidth}
    \centering
    \includegraphics[width=1.0\linewidth]{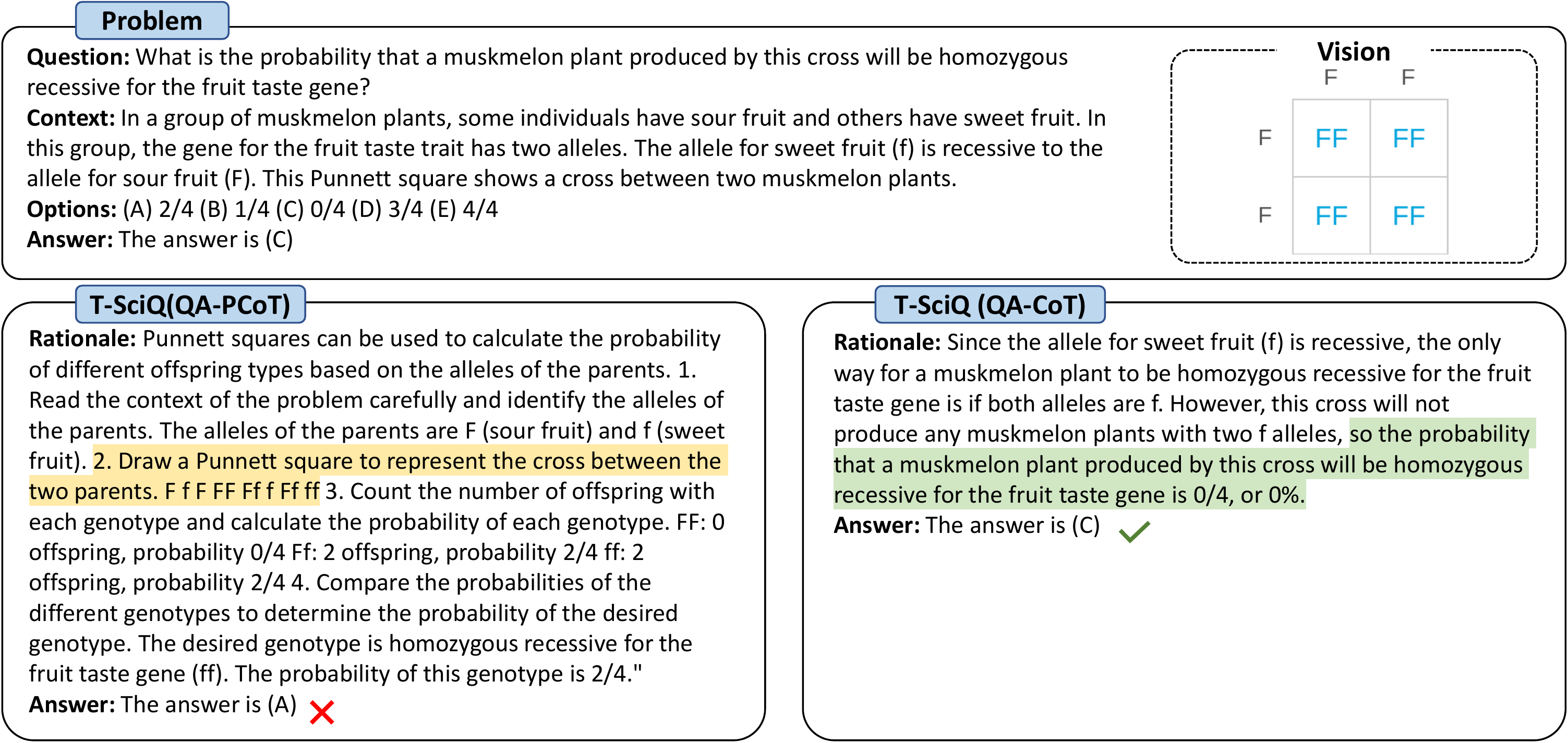}
    \caption{
       T-SciQ (QA-PCoT) may generate error rationale when visual features are complicated.
    }
    \label{fig:case_study_supp_qar_ps1}
    \end{subfigure}
    \\
    \begin{subfigure}{0.992\textwidth}
    \centering
    \includegraphics[width=1.0\linewidth]{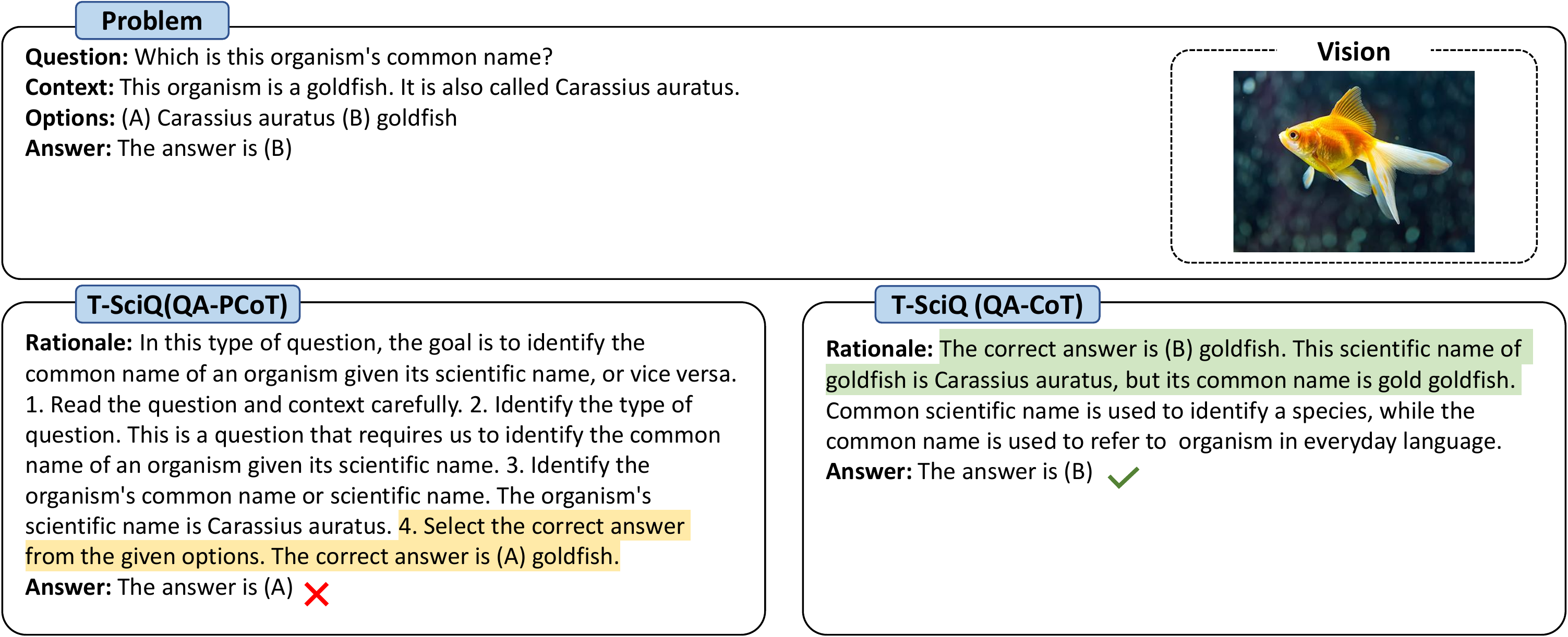}
    \caption{
        T-SciQ (QA-CoT) is better than T-SciQ (QA-PCoT) on some simple problems.
    }
    \label{fig:case_study_supp_qar_ps2}
    \end{subfigure}
    \caption{
         Examples of T-SciQ (QA-CoT) (ours) and T-SciQ (QA-PCoT) (ours) for generating rationales and predicting answers.
    }
    \label{fig:case_study_supp_qar_ps}
\end{figure*}

To investigate the impact of different teaching signals, we conducted an evaluation of model predictions trained on various types of data, including both generated teaching data and manually annotated data. In particular, we compared the test examples produced by the model trained with manually annotated data (MM-CoT) and the QA-CoT teaching data (T-SciQ (QA-CoT)) in Figure~\ref{fig:case_study_supp_qar_baseline}. We also displayed prediction cases of models trained on manually annotated data (MM-CoT) and QA-PCoT teaching data (T-SciQ (QA-PCoT)) in Figure~\ref{fig:case_study_supp_ps_baseline}.
Furthermore, we compared the prediction cases of models trained using our proposed teaching data, QA-CoT and QA-PCoT, as depicted in Figure~\ref{fig:case_study_supp_ps_qar} and Figure~\ref{fig:case_study_supp_qar_ps}. From the figures, it is evident that the model trained with QA-PCoT data performs better in solving multi-step reasoning problems, while the one trained with QA-CoT is more proficient at solving straightforward problems.

\section{Data Generation Process of PCoT}
In this section, we several show cases of how to generate PCoT teaching signals. 
To obtain appropriate planning-based chain-of-thought rationale, we introduce a 3-step zero-shot prompting approach that enables Language Models to decompose complex problems into simpler subproblems that are easier to solve. 

The lecture template used to generate a lecture for a particular skill is formulated as follows: ``Skill: \texttt{[$S$]}. QA pairs: \texttt{[$X_q, A$]} ... \texttt{[$Instruct$]}.'' In this prompt, \texttt{[$Instruct$]} is as follows: ``\textit{based on the problems above, please give a general lecture on the \texttt{[$S$]} type of question in one sentence.}''.
The lecture generated for this skill will be used in the second step of prompting, facilitating the LLM's generation of a plan for this skill.
Examples of lecture generation are shown in Table~\ref{tab:appendix-lecture-generated}.

The template used to generate a plan for a specific skill based on a lecture is formulated as follows: ``Skill: \texttt{[$S$]}. Lecture: \texttt{[$L$]}. QA pairs: \texttt{[$X_q, A$]} ... \texttt{[$Instruct$]}.''. In this prompt, \texttt{[$Instruct$]} is written as follows: ``\textit{Based on the lecture above and these problems, let’s understand these problems and devise a general and brief plan step by step to solve these problems (begin with 1, 2, 3...)}''.
The plan generated using this template will be utilized in the third stage of prompting to assist the LLM in constructing a plan-based chain-of-thought rationale for each training example.
Examples are shown in Table~\ref{tab:appendix-plan-generated}.

The lecture and plan generated by the first two prompts are used to generate a plan-based chain-of-thought rationale for each training example. The rationale generation template is formulated as follows:
``Skill: \texttt{[$S$]}. Lecture: \texttt{[$L$]}. Plan: \texttt{[$P$]}. QA pairs: \texttt{[$X_q, A$]} ... \texttt{[$Instruct$]}.''. 
In this prompt, \texttt{[$Instruct$]} is written as follows: ``Based on the lecture, the plan and the problem, please carry out the plan and solve the problem step by step (begin with 1, 2, 3...)''.
This prompting method provides a strategy for problem-solving by leveraging the expertise of the lecture and plan, and utilizing a structured reasoning framework to solve each problem.
Examples are shown in Table~\ref{tab:appendix-ps-generated}.

\begin{table*}[h]
    \centering
    \small
    \caption{Data generation process of CoT prompting.}
    \begin{tabular}{p{\linewidth}}
        \toprule
        \textbf{Input:} Which of these states is farthest north?\\
        \textbf{Context:} N/A\\
        \textbf{Options:} (A) West Virginia (B) Louisiana (C) Arizona (D) Oklahoma\\
        \textbf{Correct Answer:} (A) West Virginia \\
        \vspace{-1mm}
        Please give a detailed explanation. \\
        \sethlcolor{ppurple}
        \vspace{-1mm}
        \textbf{Solution:} \hl{West Virginia is the farthest north of the four states listed. West Virginia is located in the Appalachian region of the United States, which is in the northeastern part of the country. Louisiana, Arizona, and Oklahoma are all located in the southern and southwestern parts of the United States. West Virginia is the northernmost of the four states, making it the farthest north.} \\
        \midrule

        \textbf{Question:} Identify the question that Tom and Justin's experiment can best answer.\\
        \textbf{Context:} The passage below describes an experiment. ... They compared the distances the balls traveled when launched from a 45\degree \thinspace angle to the distances the balls traveled when launched from a 30\degree \thinspace  angle. Figure: a catapult for launching ping pong balls.\\
        \textbf{Options:} (A) Do ping pong balls stop rolling along the ground sooner after being 
        launched from a 30\degree 
        \thinspace angle or a 45\degree \thinspace angle? (B) Do ping pong balls travel farther when launched from a 30\degree \thinspace angle compared to a 45\degree \thinspace angle? \\
        \textbf{Correct Answer:} (B) Do ping pong balls travel farther when launched from a 30\degree \thinspace angle compared to a 45\degree \thinspace angle? \\
        \vspace{-1mm}
        Please give a detailed explanation. \\
        \sethlcolor{ppurple}
        \vspace{-1mm}
        \textbf{Solution:} \hl{The question that Tom and Justin's experiment can best answer is (B) Do ping pong balls travel farther when launched from a 30\degree \thinspace angle compared to a 45\degree \thinspace angle? This is because the experiment is designed to compare the distances the balls traveled when launched from a 45\degree \thinspace angle to the distances the balls traveled when launched from a 30\degree \thinspace angle. By measuring the distances between the catapult and the place where the ball hit the ground, Tom and Justin can determine which angle produces the greatest distance.} \\

        \midrule

        \textbf{Question:} What does the verbal irony in this text suggest? According to Mr. Herrera's kids, his snoring is as quiet as a jackhammer.\\
        \textbf{Context:} N/A\\
        \textbf{Options:} (A) The snoring is loud. (B) The snoring occurs in bursts. \\
        \textbf{Correct Answer:} (A) The snoring is loud. \\
        \vspace{-1mm}
        Please give a detailed explanation. \\
        \sethlcolor{ppurple}
        \vspace{-1mm}
        \textbf{Solution:} \hl{Verbal irony is when someone says something that is the opposite of what they mean. In this text, Mr. Herrera's kids are saying that his snoring is as quiet as a jackhammer, which implies that it is actually very loud. This suggests that the snoring is loud, which is option (A).} \\

        \midrule

        \textbf{Question:} Which of the following could Wendy's test show? \\
        \textbf{Context:} People can use the engineering-design process to develop solutions to problems. One step in the process is testing if a potential solution meets the requirements of the design. ... She read that giving the bacteria more nutrients could affect the amount of insulin they produced. So, Wendy gave extra nutrients to some of the bacteria. Then, she measured how much insulin those bacteria produced compared to bacteria that did not get extra nutrients. Figure: studying bacteria in a laboratory. \\
        \textbf{Options:} (A) whether producing more insulin would help the bacteria grow faster (B) whether different types of bacteria would need different nutrients to produce insulin (C) whether she added enough nutrients to help the bacteria produce 20\% more insulin \\
        \textbf{Correct Answer:} (C) whether she added enough nutrients to help the bacteria produce 20\% more insulin \\
        \vspace{-1mm}
        Please give a detailed explanation. \\
        \sethlcolor{ppurple}
        \vspace{-1mm}
        \textbf{Solution:} \hl{The passage describes how Wendy used the engineering-design process to test a solution to a problem. She wanted to increase the amount of insulin that the bacteria produced by 20\%, so she gave extra nutrients to some of the bacteria and then measured how much insulin those bacteria produced compared to bacteria that did not get extra nutrients. This testing would show whether she added enough nutrients to help the bacteria produce 20\% more insulin. Option A is incorrect because the passage does not mention anything about the bacteria growing faster. Option B is incorrect because the passage does not mention anything about different types of bacteria needing different nutrients.} \\

        \midrule

        \textbf{Question:} Does this passage describe the weather or the climate? \\
        \textbf{Context:} Figure: Chicago. Chicago is known as The Windy City. But on average, the wind there only blows at about 10 miles per hour. Hint: Weather is what the atmosphere is like at a certain place and time. Climate is the pattern of weather in a certain place.\\
        \textbf{Options:} (A) weather (B) climate \\
        \textbf{Correct Answer:} (B) climate \\
        \vspace{-1mm}
        Please give a detailed explanation. \\
        \sethlcolor{ppurple}
        \vspace{-1mm}
        \textbf{Solution:} \hl{The passage describes the average wind speed in Chicago, which is a pattern of weather in the city. This pattern of weather is part of the climate of Chicago, so the correct answer is (B) climate. Climate is the average weather conditions in a certain area over a long period of time. It is determined by factors such as temperature, precipitation, humidity, and wind speed. Weather, on the other hand, is the atmospheric conditions at a certain place and time. It is determined by the same factors as climate, but it is more localized and can change quickly.} \\
        
        \bottomrule

    \end{tabular}
    \label{tab:appendix-qar-generated}
\end{table*}

\begin{table*}[h]
    \centering
    \small
    \caption{The process of generating lectures.}
    \begin{tabular}{p{\linewidth}}
        \toprule
        Here are some problems about "Read a map: cardinal directions" \\

        \vspace{-1mm}
        
        \textbf{Question:} Which of these states is farthest north? \\
        \textbf{Context:} N/A \\
        \textbf{Options:} (A) West Virginia (B) Louisiana (C) Arizona (D) Oklahoma \\
    
        \textbf{Question:} Which of these states is farthest north? \\
        \textbf{Context:} N/A \\
        \textbf{Options:} (A) Alabama (B) Idaho (C) South Carolina (D) Oklahoma \\

        \textbf{Question:} Which of these states is farthest north? \\
        \textbf{Context:} N/A \\
        \textbf{Options:} (A) Maine (B) South Carolina (C) Kansas (D) Delaware \\
        
        \textbf{Question:} Which of these states is farthest east? \\
        \textbf{Context:} N/A \\
        \textbf{Options:} (A) Montana (B) New Mexico (C) Arkansas (D) Nebraska \\
        
        \textbf{Question:} Which of these states is farthest north? \\
        \textbf{Context:} N/A \\
        \textbf{Options:} (A) West Virginia (B) Mississippi (C) Nebraska (D) Oklahoma \\
        
        \vspace{-1mm}
        
        Based on the problems above, please give a general lecture on the "Read a map: cardinal directions" type of question in one sentence. \\
        
        \sethlcolor{ppurple}
        \vspace{-1mm}
        \textbf{Lecture:} \hl{Reading a map to identify cardinal directions involves determining which direction a given location is in relation to other locations on the map.} \\
        \midrule

        Here are some problems about "Interpret figures of speech" \\

        \vspace{-1mm}

        \textbf{Question:} What does the verbal irony in this text suggest?According to Mr. Herrera's kids, his snoring is as quiet as a jackhammer. \\
        \textbf{Context:} N/A \\
        \textbf{Options:} (A) The snoring is loud. (B) The snoring occurs in bursts. \\
        
        \textbf{Question:} What does the metaphor in this text suggest? All religions, arts, and sciences are branches of the same tree. \u2014Albert Einstein \\
        \textbf{Context:} N/A \\
        \textbf{Options:} (A) All religions, arts, and sciences are distant from one another. (B) All religions, arts, and sciences are related. \\
        
        \textbf{Question:} What does the verbal irony in this text suggest? This is the best day of my life, Mr. Hogan mumbled after his car broke down on the way to an important job interview. \\
        \textbf{Context:} N/A \\
        \textbf{Options:} (A) Mr. Hogan was having a bad day. (B) Mr. Hogan liked fixing cars. \\
        
        \textbf{Question:} What does the allusion in this text suggest?Tyler seems to have the Midas touch. Without any special experience or training, he launched a thriving business and then established a well-respected charity. \\
        \textbf{Context:} N/A \\
        \textbf{Options:} (A) Tyler is successful at all that he does. (B) Tyler has a hands-on approach to his work. \\
        
        \textbf{Question:} What does the idiom in this text suggest? Speak of the devil! Whitney declared when Charlie strolled into the room. \\
        \textbf{Context:} N/A \\
        \textbf{Options:} (A) Whitney had just been speaking about Charlie. (B) Whitney didn't trust Charlie. \\
    
        \vspace{-1mm}

        Based on the problems above, please give a general lecture on the "Interpret figures of speech" type of question in one sentence. \\

        \sethlcolor{ppurple}
        \vspace{-1mm}
        \textbf{Lecture:} \hl{Interpret figures of speech questions require the reader to identify the figurative language used in a text and determine the meaning or connotation it conveys.} \\
            
        \midrule

        Here are some problems about "Animal adaptations: beaks, mouths, and necks" \\

        \vspace{-1mm}

        \textbf{Question:} Which animal's mouth is also adapted for bottom feeding? \\
        \textbf{Context:} Sturgeons eat invertebrates, plants, and small fish. They are bottom feeders. Bottom feeders find their food at the bottom of rivers, lakes, and the ocean. The 's mouth is located on the underside of its head and points downward. Its mouth is adapted for bottom feeding. Figure: sturgeon.  \\
        \textbf{Options:} (A) discus (B) armored catfish \\
        
        \textbf{Question:} Which bird's beak is also adapted to tear through meat? \\
        \textbf{Context:} Red-tailed hawks eat fish, mammals, and other birds. The shape of the 's beak is adapted to tear through meat. Figure: red-tailed hawk. \\
        \textbf{Options:} (A) sand martin (B) Cape vulture \\
        
        \textbf{Question:} Which animal's mouth is also adapted for bottom feeding? \\
        \textbf{Context:} Armored catfish eat plants and small \\ invertebrates. They are bottom feeders. Bottom feeders find their food at the bottom of rivers, lakes, and the ocean. The catfish's mouth is located on the underside of its head and points downward. Its mouth is adapted for bottom feeding. Figure: armored catfish. \\
        \textbf{Options:} (A) clown triggerfish (B) sturgeon \\

        \textbf{Question:} Which bird's beak is also adapted to get nectar out of long flowers? \\
        \textbf{Context:} Green violetears live in the forests of Central and South America. The shape of the 's beak is adapted to get nectar out of long flowers. Figure: green violetear. \\
        \textbf{Options:} (A) ground hornbill (B) violet sabrewing \\

        \textbf{Question:} ... \\

        \vspace{-1mm}

        Based on the problems above, please give a general lecture on the "Animal adaptations: beaks, mouths, and necks" type of question in one sentence. \\

        \sethlcolor{ppurple}
        \vspace{-1mm}
        \textbf{Lecture:} \hl{Animal adaptations such as beaks, mouths, and necks are specialized features that enable animals to feed on different types of food, such as meat, insects, nuts, and plant matter.} \\
        
        \bottomrule

    \end{tabular}
    \label{tab:appendix-lecture-generated}
\end{table*}

\begin{table*}[h]
    \centering
    \small
    \caption{The process of generating plans.}
    \begin{tabular}{p{\linewidth}}
        \toprule
        Here are some problems about "Read a map: cardinal directions" \\

        \vspace{-1mm}

        The lecture about "Read a map: cardinal directions" is "Reading a map to identify cardinal directions involves determining which direction a given location is in relation to other locations on the map." \\

        \vspace{-1mm}
        
        \textbf{Question:} Which of these states is farthest north? \\
        \textbf{Context:} N/A \\
        \textbf{Options:} (A) West Virginia (B) Louisiana (C) Arizona (D) Oklahoma \\
    
        \textbf{Question:} Which of these states is farthest north? \\
        \textbf{Context:} N/A \\
        \textbf{Options:} (A) Alabama (B) Idaho (C) South Carolina (D) Oklahoma \\

        \textbf{Question:} Which of these states is farthest north? \\
        \textbf{Context:} N/A \\
        \textbf{Options:} (A) Maine (B) South Carolina (C) Kansas (D) Delaware \\

        \textbf{Question:} ... \\
        
        
        
        \vspace{-1mm}
        
        Based on the lecture above and these problems, let's understand these problems and devise a general and brief plan step by step to solve these problems (begin with 1, 2, 3...). \\
        
        \sethlcolor{ppurple}
        \vspace{-1mm}
        \textbf{Plan:} \hl{1. Read the question carefully and identify the context. 2. Identify the cardinal directions (north, south, east, and west) in relation to the given locations on the map. 3. Compare the locations and determine which one is farthest in the given direction. 4. Select the correct answer from the given options.} \\
        \midrule

        Here are some problems about "Interpret figures of speech" \\

        \vspace{-1mm}

        The lecture about "Interpret figures of speech" is "Interpret figures of speech questions require the reader to identify the figurative language used in a text and determine the meaning or connotation it conveys." \\

        \vspace{-1mm}

        \textbf{Question:} What does the verbal irony in this text suggest?According to Mr. Herrera's kids, his snoring is as quiet as a jackhammer. \\
        \textbf{Context:} N/A \\
        \textbf{Options:} (A) The snoring is loud. (B) The snoring occurs in bursts. \\
        
        \textbf{Question:} What does the metaphor in this text suggest? All religions, arts, and sciences are branches of the same tree. \u2014Albert Einstein \\
        \textbf{Context:} N/A \\
        \textbf{Options:} (A) All religions, arts, and sciences are distant from one another. (B) All religions, arts, and sciences are related. \\

        \textbf{Question:} What does the verbal irony in this text suggest? This is the best day of my life, Mr. Hogan mumbled after his car broke down on the way to an important job interview. \\
        \textbf{Context:} N/A \\
        \textbf{Options:} (A) Mr. Hogan was having a bad day. (B) Mr. Hogan liked fixing cars. \\

        \textbf{Question:} ... \\
        
        
    
        \vspace{-1mm}

        Based on the lecture above and these problems, let's understand these problems and devise a general and brief plan step by step to solve these problems (begin with 1, 2, 3...). \\

        \sethlcolor{ppurple}
        \vspace{-1mm}
        \textbf{Plan:} \hl{1. Read the text carefully and identify the figure of speech used. 2. Analyze the context of the text to determine the meaning or connotation of the figure of speech. 3. Choose the option that best matches the meaning or connotation of the figure of speech.} \\
            
        \midrule

        Here are some problems about "Animal adaptations: beaks, mouths, and necks" \\

        \vspace{-1mm}

        The lecture about "Animal adaptations: beaks, mouths, and necks" is "Animal adaptations such as beaks, mouths, and necks are specialized features that enable animals to feed on different types of food, such as meat, insects, nuts, and plant matter." \\

        \vspace{-1mm}

        \textbf{Question:} Which animal's mouth is also adapted for bottom feeding? \\
        \textbf{Context:} Sturgeons eat invertebrates, plants, and small fish. They are bottom feeders. Bottom feeders find their food at the bottom of rivers, lakes, and the ocean. The 's mouth is located on the underside of its head and points downward. Its mouth is adapted for bottom feeding. Figure: sturgeon.  \\
        \textbf{Options:} (A) discus (B) armored catfish \\
        
        \textbf{Question:} Which bird's beak is also adapted to tear through meat? \\
        \textbf{Context:} Red-tailed hawks eat fish, mammals, and other birds. The shape of the 's beak is adapted to tear through meat. Figure: red-tailed hawk. \\
        \textbf{Options:} (A) sand martin (B) Cape vulture \\
        
        \textbf{Question:} Which animal's mouth is also adapted for bottom feeding? \\
        \textbf{Context:} Armored catfish eat plants and small \\ invertebrates. They are bottom feeders. Bottom feeders find their food at the bottom of rivers, lakes, and the ocean. The catfish's mouth is located on the underside of its head and points downward. Its mouth is adapted for bottom feeding. Figure: armored catfish. \\
        \textbf{Options:} (A) clown triggerfish (B) sturgeon \\


        \textbf{Question:} ... \\

        \vspace{-1mm}

        Based on the lecture above and these problems, let's understand these problems and devise a general and brief plan step by step to solve these problems (begin with 1, 2, 3...). \\

        \sethlcolor{ppurple}
        \vspace{-1mm}
        \textbf{Plan:} \hl{1. Read the lecture and understand the context of the questions. 2. Read the question and identify the type of adaptation being asked about. 3. Look at the figure provided and identify the animal being discussed. 4. Research the animal to find out what type of adaptation it has. 5. Compare the options provided and select the one that best matches the adaptation.} \\
        
        \bottomrule

    \end{tabular}
    \label{tab:appendix-plan-generated}
\end{table*}

\begin{table*}[h]
    \centering
    \small
    \caption{Data generation of PCoT prompting.}
    \begin{tabular}{p{\linewidth}}
        \toprule
        Here are is a problem about "Read a map: cardinal directions" \\

        \vspace{-1mm}

        The lecture about "Read a map: cardinal directions" is "Reading a map to identify cardinal directions involves determining which direction a given location is in relation to other locations on the map." \\

        \vspace{-1mm}

        The plan to solve "Read a map: cardinal directions" problem is "1. Read the question carefully and identify the context. 2. Identify the cardinal directions (north, south, east, and west) in relation to the given locations on the map. 3. Compare the locations and determine which one is farthest in the given direction. 4. Select the correct answer from the given options." \\

        \vspace{-1mm}
        
        \textbf{Question:} Which of these states is farthest north? \\
        \textbf{Context:} N/A \\
        \textbf{Options:} (A) West Virginia (B) Louisiana (C) Arizona (D) Oklahoma \\
        \textbf{Correct Answer:} (A) West Virginia \\
        
        \vspace{-1mm}
        
        Based on the lecture, the plan and the problem, please carry out the plan and solve the problem step by step (begin with 1, 2, 3...). \\
        
        \sethlcolor{ppurple}
        \vspace{-1mm}
        \textbf{Solution:} \hl{1. Read the question carefully and identify the context. The context of the question is to determine which of the given states is farthest north. 2. Identify the cardinal directions (north, south, east, and west) in relation to the given locations on the map. North is the direction that is farthest away from the equator and closest to the North Pole. South is the direction that is farthest away from the North Pole and closest to the equator. East is the direction that is farthest away from the Prime Meridian and closest to the International Date Line. West is the direction that is farthest away from the International Date Line and closest to the Prime Meridian. 3. Compare the locations and determine which one is farthest in the given direction. West Virginia is farthest north, followed by Louisiana, Arizona, and Oklahoma. 4. Select the correct answer from the given options. The correct answer is (A) West Virginia.} \\
        \midrule

        Here is a problem about "Interpret figures of speech" \\

        \vspace{-1mm}

        The lecture about "Interpret figures of speech" is "Interpret figures of speech questions require the reader to identify the figurative language used in a text and determine the meaning or connotation it conveys." \\

        \vspace{-1mm}

        The plan to solve "Interpret figures of speech" problem is "1. Read the text carefully and identify the figure of speech used. 2. Analyze the context of the text to determine the meaning or connotation of the figure of speech. 3. Choose the option that best matches the meaning or connotation of the figure of speech." \\

        \vspace{-1mm}

        \textbf{Question:} What does the verbal irony in this text suggest?According to Mr. Herrera's kids, his snoring is as quiet as a jackhammer. \\
        \textbf{Context:} N/A \\
        \textbf{Options:} (A) The snoring is loud. (B) The snoring occurs in bursts. \\
        \textbf{Correct Answer:} (A) The snoring is loud. \\
        
        \vspace{-1mm}

        Based on the lecture, the plan and the problem, please carry out the plan and solve the problem step by step (begin with 1, 2, 3...). \\

        \sethlcolor{ppurple}
        \vspace{-1mm}
        \textbf{Solution:} \hl{1. Read the text carefully and identify the figure of speech used. The figure of speech used in the text is verbal irony. 2. Analyze the context of the text to determine the meaning or connotation of the figure of speech. The context of the text suggests that the verbal irony is used to convey the idea that Mr. Herrera's snoring is actually very loud, despite what his kids say. 3. Choose the option that best matches the meaning or connotation of the figure of speech. The option that best matches the meaning or connotation of the figure of speech is (A) The snoring is loud.} \\
            
        \midrule

        Here is a problem about "Animal adaptations: beaks, mouths, and necks" \\

        \vspace{-1mm}

        The lecture about "Animal adaptations: beaks, mouths, and necks" is "Animal adaptations such as beaks, mouths, and necks are specialized features that enable animals to feed on different types of food, such as meat, insects, nuts, and plant matter." \\

        \vspace{-1mm}

        The plan to solve "Animal adaptations: beaks, mouths, and necks" problem is "1. Read the lecture and understand the context of the questions. 2. Read the question and identify the type of adaptation being asked about. 3. Look at the figure provided and identify the animal being discussed. 4. Research the animal to find out what type of adaptation it has. 5. Compare the options provided and select the one that best matches the adaptation." \\

        \textbf{Question:} Which animal's mouth is also adapted for bottom feeding? \\
        \textbf{Context:} Sturgeons eat invertebrates, plants, and small fish. They are bottom feeders. Bottom feeders find their food at the bottom of rivers, lakes, and the ocean. The 's mouth is located on the underside of its head and points downward. Its mouth is adapted for bottom feeding. Figure: sturgeon.  \\
        \textbf{Options:} (A) discus (B) armored catfish \\
        \textbf{Correct Answer:} (B) armored catfish \\
        
        \vspace{-1mm}

        Based on the lecture, the plan and the problem, please carry out the plan and solve the problem step by step (begin with 1, 2, 3...). \\

        \sethlcolor{ppurple}
        \vspace{-1mm}
        \textbf{Solution:} \hl{1. Read the lecture and understand the context of the questions. The lecture states that animal adaptations such as beaks, mouths, and necks are specialized features that enable animals to feed on different types of food. 2. Read the question and identify the type of adaptation being asked about. The question is asking about an adaptation related to bottom feeding. 3. Look at the figure provided and identify the animal being discussed. The figure provided is a sturgeon. 4. Research the animal to find out what type of adaptation it has. Researching the sturgeon reveals that its mouth is located on the underside of its head and points downward. This adaptation is specifically adapted for bottom feeding. 5. Compare the options provided and select the one that best matches the adaptation. The options provided are (A) discus and (B) armored catfish. The armored catfish has a mouth that is adapted for bottom feeding, making it the correct answer. Therefore, the correct answer is (B) armored catfish.} \\
        
        \bottomrule

    \end{tabular}
    \label{tab:appendix-ps-generated}
\end{table*}

\end{document}